\DocumentMetadata{}

\documentclass[screen]{acmart}

\AtBeginDocument{%
  }

\setcopyright{acmcopyright}
\copyrightyear{2024}
\acmYear{2024}
\acmDOI{XXXXXXX.XXXXXXX}

\acmISBN{978-1-4503-XXXX-X/18/06}

\usepackage[edges]{forest}
\usepackage{fix-cm}

\usepackage[figuresright]{rotating}
\usepackage{adjustbox,cleveref,graphicx,soul,}
\crefname{figure}{Fig.}{Figs.}
\crefname{section}{Section}{Sections}
\crefname{table}{Table}{Tables}
\usepackage[inline]{enumitem}
\usepackage[justification=centering]{caption}
\usepackage{booktabs,array,longtable,rotating}
\usepackage{multirow}

\usepackage{expl3}
\usepackage{tabularray}
\UseTblrLibrary{varwidth}  
\UseTblrLibrary{booktabs}

\DeclareRobustCommand{\hlcyan}[1]{{\sethlcolor{cyan}\hl{#1}}}
\DeclareRobustCommand{\hlblue}[1]{{\sethlcolor{magenta}\hl{#1}}}
\DeclareRobustCommand{\hlred}[1]{{\sethlcolor{pink}\hl{#1}}}
\DeclareRobustCommand{\hlorange}[1]{{\sethlcolor{orange}\hl{#1}}}
\DeclareRobustCommand{\hlpurple}[1]{{\sethlcolor{magenta}\hl{#1}}}
\DeclareRobustCommand{\hlgreen}[1]{{\sethlcolor{lime}\hl{#1}}}

\usepackage[bottom]{footmisc}
\usetikzlibrary{shapes.geometric,decorations.text,shadows,calc,shadows.blur}

\def\aSlider#1#2{
	\tikz[baseline=-0.1cm]{
		\coordinate (start) at (0,0);
		\coordinate (end) at (#1,0);
		\coordinate (mark) at ($(start)!#2!(end)$);
		\useasboundingbox (start|- 0,-.25) rectangle (end|- 0, .25);
		\draw[line width=0.4mm, line cap=round, blue!50!cyan] 
		(start) -- (mark) edge[lightgray] (end);
		\node[fill=white, draw=lightgray, very thin,
		blur shadow={shadow xshift=0pt, shadow opacity=20, shadow yshift=-0.9mm,
			shadow blur steps=6, shadow blur radius=0.3mm},
		circle, minimum size=0.25cm, inner sep=0pt] at (mark) {};
	}
}
\makeatletter
\newcommand*{\radiobutton}{%
	\@ifstar{\@radiobutton0}{\@radiobutton1}%
}
\newcommand*{\@radiobutton}[1]{%
	\begin{tikzpicture}
		\pgfmathsetlengthmacro\radius{height("X")/2}
		\draw[radius=\radius] circle;
		\ifcase#1 \fill[radius=.6*\radius] circle;\fi
	\end{tikzpicture}%
}
\makeatother

\usepackage[many]{tcolorbox}
\newenvironment{example}[1]{%
	\tcolorbox[fontlower=\footnotesize\itshape,fontupper=\footnotesize,fonttitle=\footnotesize\bfseries,coltitle=black,arc=0mm,colbacktitle=gray!15,colback=gray!15,
	top=0mm,toptitle=.5em,
	middle=0mm,boxrule=.0mm,title=#1,fontlower=\footnotesize,]
}{\endtcolorbox}

\colorlet{col1}{white}
\colorlet{col2}{lime!20}
\colorlet{col3}{pink!50}
\forestset{ 
	typology/.style={
		for tree={grow=east,forked edges,draw,
			calign=last,
			anchor=base
			west, tier/.pgfmath=level(),
			fork sep=5pt,drop shadow,
			,},
		where level=0{top color=col3,bottom color=col3,	draw=black}{},
		where level=1{top color=col2,bottom color=col2,	draw=black}{},
		where level=2{top color=col1,bottom color=col1,	draw=black}{}, 	
		where level=3{top color=col1,bottom color=col1,	draw=black}{}, 
	},
}

\newcommand{\dimension}[1]{{\sffamily#1}}

\usepackage{listings}
\usepackage[scaled=0.9]{beramono}
\usepackage{microtype}
\newcommand\Smallish{\fontsize{8}{10}\selectfont}
\newcommand*\LSTfont{%
	\Smallish\ttfamily\SetTracking{encoding=*}{-50}\lsstyle}

\usepackage{scalefnt}
\newcommand\extrasmall[2][0.85]{{\scalefont{#1}#2}}
\newcommand{\req}[1]{(\extrasmall{RQ#1})}

\begin{document}

\title[A typology of Explanations for Explainability-by-Design]{A Typology of Explanations for Explainability-by-Design}

\author{Niko Tsakalakis}
\orcid{0000-0003-2654-0825}
\email{nk.ts@icloud.com}

\author{Sophie Stalla-Bourdillon}
\orcid{0000-0003-3715-1219}
\affiliation{%
  \institution{Vrije Universiteit Brussel}
  \city{Brussels}
  \country{Belgium}
}
\email{sophie.stalla-bourdillon@vub.be}

\author{Trung Dong Huynh}
\orcid{0000-0003-4937-2473}
\affiliation{%
	\institution{King's College London}
	\city{London}
	\country{United Kingdom}
}
\email{dong.huynh@kcl.ac.uk}

\author{Luc Moreau}
\orcid{0000-0002-3494-120X}
\affiliation{%
	\institution{University of Sussex}
	\city{Falmer}
	\country{United Kingdom}
}
\email{luc.moreau@sussex.ac.uk}

\renewcommand{\shortauthors}{Tsakalakis et al.}

\begin{abstract}
As automated decision-making permeates almost all aspects of everyday life, capabilities to generate meaningful explanations for various stakeholders (i.e., decision-makers, addressees of decisions including individuals, auditors, and regulators) should be carefully deployed. This paper presents a typology of explanations intended to support the first pillar of an explainability-by-design strategy. Its production has been achieved by pursuing a responsible innovation approach and introducing a new persona within the research and innovation process, i.e., a legal engineer, whose role is to work at the interface of two teams, the compliance and the engineering teams and to oversee the process of requirement elicitation, which is often opinionated and narrowing. Once explanation requirements have been derived from applicable regulatory requirements, compliance rules or business policies, they have been mapped to the dimensions of the typology to produce fine-grained explanation requirements, forming computable building blocks that can then be translated into system requirements during the technical design phase. The typology has been co-created with industry partners operating in two sectors: finance and education. Two pilot studies have thus been conducted to test both the feasibility of the generation and computation of explanations on the basis of the typology and the usefulness of the outputs in the light of the state of the art. The typology comprises nine hierarchical dimensions. It can be leveraged to operate a stand-alone classifier of explanations that acts as detective controls within a broader partially-automated compliance strategy. A machine-readable format of the typology is provided in the form of a light ontology.
\end{abstract}



\received{11 December 2023}
\received[revised]{16 September 2024}
\received[accepted]{12 November 2024}

\maketitle

\section{Introduction}\label{section:introduction}

Automated decisions and the ways in which they can be interpreted by and explained to humans have been at the forefront of Artificial Intelligence (AI) research for a while now.\negmedspace\footnote{See e.g.\ for the evolution of interpretable tools and the emergence of Explainable AI (XAI)~\cite{Preece:2018,Arrieta:2020,Liao2021}.} Explainability has been described as being a powerful tool for a variety of purposes including ``detecting flaws in the model and biases in the data, for verifying predictions, for improving models, and finally for gaining new insights into the problem at hand''~\cite{Samek2017}. However, there is no consensus on the metrics for evaluating explanation methods, and it is hard to compare them~\cite{Zhou:2021}. \citeauthor{Kim:2022} argue that explainable AI cannot limit itself to assessing the simplicity and locality of an explanation and  social, ethical, and epistemic values should be taken into account to assert the relevance of a range of explanation types and stakeholders in explaining AI outputs~\citep{Kim:2022}. More recently, it has been argued that ``without a proper theory of idealization, it remains difficult to thoroughly diagnose the success of xAI methods''~\cite{Sullivan2024}. 

Understanding how and why automated decision-making systems process personal data of individuals became particularly important with the introduction of the General Data Protection Regulation (GDPR)~\cite{GDPR} in the European Union (EU). Its Article 22 on `\textit{Automated individual decision-making, including profiling}'
gave rise to a debate about the existence of a right to explanation.\negmedspace\footnote{See e.g.~\cite{Selbst:2017,Edwards:2017,Wachter:2017}.} National data protection authorities' and courts' decisions have confirmed the existence of a right to explanation, although its contours are not clearly delimited yet~\cite{BarrosVale:2022, Ausloos:2025}. In addition, the Court of Justice of the European Union (CJEU) has adopted a broad definition of automated individual decision-making, which includes ``the automated establishment, by a credit information agency, of a probability value based on personal data relating to a person and concerning his or her ability to meet payment commitments in the future (\ldots), where a third party, to which that probability value is transmitted, draws strongly on that probability value to establish, implement or terminate a contractual relationship with that person.\negmedspace\footnote{CJEU C‑634/21 OQ v Land Hessen 7 December 2023 ECLI:EU:C:2023:957, para. 73}''
Regardless of the contours and source of such a right,\negmedspace\footnote{Opinion of Advocate General Richard de la Tour C‑203/22 CK 12 September 2024 ECLI:EU:C:2024:745, para. 67 (``In short, `meaningful information', as required by Article 15(1)(h) of the GDPR, must not only be clear and accessible, but must also be accompanied by explanations to ensure that it is properly understood'').} the debate provided an opportunity to shift the focus from the inner workings of the `black box',
which has been the primary focus of XAI,
to its broader environment and purpose, i.e., the production of information meaningful for the exercise of individual rights
\citep{Selbst:2017,Cobbe:2020}. 
Recent discussions on explanations have highlighted the limits of existing approaches for generating 
meaningful explanations, e.g.\ the lack of focus upon user needs~\cite{Yang2021}, the difference between explanations that offer a justification (they respond to the `why' question) or produce transparency (they respond to the `how' question) \citep{Pieters:2010}, with some indication that end users tend to pay more attention to the `why' question~\cite{Chazette2019}, as confirmed by an overview of research on explanations in philosophy, cognitive science, and
the social sciences. \citeauthor{Mittelstadt2018} thus argue that explanations should be ``contrastive, selective, and socially interactive''~\cite{Mittelstadt2018}.
At the same time, judges are increasingly stressing the importance of contextualising the personal data processed to ensure that the information received by data subjects is intelligible.\negmedspace\footnote{CJEU Case C-487/21 FF v Österreichische Datenschutzbehörde and CRIF GmbH, 4 May 2023 ECLI:EU:C:2023:369, para. 41), and CJEU C‑307/22 FT v DW 26 October 2023 ECLI:EU:C:2023:811, para. 74.}

Several commentators focus upon specific groups of consumers of explanations, i.e., data subjects~\cite{Malgieri:2017}, developers~\cite{Zhu2018} without directly addressing differing levels of understanding. Often, proposed solutions have attempted to explain models while ignoring the impact of factors that are external to black boxes~\cite{Cobbe:2020,Edwards:2017}, 
or have not taken into account events occurring before the training stage or during model deployment, with a few exceptions like~\cite{Huynh2021,Cobbe:2019}.
One type of explanations in particular, i.e., counterfactual explanations~\cite{Wachter:2017a},
has been gaining traction in practice despite its inherent limits and in particular its inability to address the `how' question.\negmedspace\footnote{By way of example, counterfactual explanations are not necessarily helpful in providing useful information to challenge a financial decision, such as a rejection of a loan application. See in general~\cite{Barocas:2019}.}

The need to develop a systematic approach to explanations for GDPR compliance has thus been discussed extensively in the literature as a high-level compliance goal\footnote{See, for example,~\cite{Martin_2018,Labadie:2019}.}
or strategy.\negmedspace\footnote{See, for example,~\cite{Deng:2010,Notario:2015}.}
With this said, rights to explanations are also present in sector-specific regulatory frameworks, such as education and in particular access to schools, as explained in Section~\ref{subsection: school allocation}, and the financial sectors, as mentioned in Section~\ref{subsection: loan application}. As regards the financial sector, it is worth mentioning that a new Directive on credit agreements for consumers has been adopted by the EU in 2023, which builds on the GDPR and further specifies the rights to information and explanation in the context of covered credit agreements~\cite{ConsumerCreditDirective}.\negmedspace\footnote{Article 18(8)(a), in particular, provides that ``[w]here the creditworthiness assessment involves the use of automated processing of personal data, Member States shall ensure that the consumer has the right to request and obtain from the creditor human intervention, consisting of the right to request and obtain from the creditor a clear and comprehensible explanation of the assessment of creditworthiness, including on the logic and risks involved in the automated processing of personal data as well as its significance and effects on the decision''.} National transpositions will be applicable from 20 November 2026. 
Horizontal frameworks, such as the EU AI Act~\cite{AIACT} can also introduce bespoke rights to explanations, as discussed in Section~\ref{subsection: AI Act}. There is now a pressing need to better understand what explanation outputs could look like in practice to advance the state of the art. 

Analysing the characteristics of explanations through the prisms of philosophy and data science, \citet{Pieters:2010} distinguishes between explanations-for-trust and explanations-for-confidence. Essentially, explanations sit on a spectrum. 
Explanations with very little detail fail, as they do not contain enough information to build recipients' confidence or trust. 
Explanations that contain too much detail fail as well because they are misaligned with the recipient's level of understanding. 
Explanations that contain a low level of detail can help build confidence by answering the `why' question, and explanations that contain a high level of detail can help build trust by answering the `how' question.
The `how' question is of interest for assessing data quality, e.g.\ to help users understand how their data contributed to algorithmic intelligence~\cite{Chazette2020}.
Pieters also notes that different explanations can be generated to answer the same question~\citep{Pieters:2010}. 

Arguing for a right to `meaningful and intelligible explanation', \citet{Kim:2022} distinguish between \textit{ex ante} generic explanations to achieve informed consent and \textit{ex post} generic and specific explanations either to inform remedial actions or to produce individualised information about the algorithmic processing.
\citet{CastetsRenard:2021}, categorising the impact levels of predictive policing decisions into four classes,\negmedspace\footnote{%
	Ranging from Level I `no impact' to Level IV `very high impact' according to the Impact Assessment Levels of the Canadian Directive on Automated Decision Making.} 
distinguishes explanation types by triggering event: for decisions with no impact on the individual, a publicly available (generic) explanation should suffice; 
for decision with a moderate impact, explanations should be produced upon a request from the user; 
for decisions with a high or very high impact, explanations should be systematically produced at the point of disclosure of the decision.

The GDPR, and the introduction of data protection by design as a mandatory requirement,\negmedspace\footnote{GDPR, Article 25.} has fed a formalisation effort and attempts to make GDPR obligations machine readable, to produce GDPR-specific description languages like Fides,\negmedspace\footnote{\href{https://ethyca.com/fides/}{https://ethyca.com/fides/}}
to create a  GDPR vocabulary for use in ontologies~\cite{Bartolini:2015}, 
or to extend a Data Privacy Vocabulary with GPDR concepts.\negmedspace\footnote{\url{http://www.w3id.org/dpv/dpv-gdpr\#}}
However, these methods have not been applied to the generation or computation of explanations yet. 
\citet{Pieters:2010}, nonetheless, acknowledges the relevance of explanations programs. 
\begin{quote}
	\textit{``Actants can thus be said to have an explanation program, i.e.\ their action program projected on the domain of explanation. When actants are asked to explain something about a theory or system, they have certain intentions and possibilities for explaining in a certain way.''}
\end{quote}
He also observes that responsibilities for explanations can be delegated, either to humans or machines. 
When responsibilities for explanations are delegated to machines, however, explanation-for-trust and explanation-for-confidence are then needed to backup computed explanations.

The purpose of this paper is to lay the foundations for building partially-automated explanation programs following a responsible innovation approach. Innovating responsibly implies a ``continuous commitment to be anticipatory, reflective, inclusively deliberative, and responsive''~\cite{Owen:2013}, which can be pursued through a variety of practices, like socio-technical integration research practices.\negmedspace\footnote{For an example see https://cta-toolbox.nl/tools/STIR/}
This paper thus intentionally departs from approaches that would leverage imprecise, probabilistic techniques, such as generative AI, to produce explanations~\cite{Pasquale:2024} and acknowledges the inherent limits of any formalisation effort when legal requirements are at stake. In fact, computed explanatory statements are conceived as a means to facilitate the task of the entity accountable for the decision or responsible for explaining the decision: they are thus merely supporting statements, which might need to be refined and completed iteratively to meet the needs of a particular addressee of the decision, such as a data subject. 

The first question to explore when building an automated explanation program relates to the building of a methodology for generating a comprehensive set of computable explanations at each relevant node of a data processing pipeline. In other words, we need a methodology for pursuing an explainability-by-design strategy, conceived as a form of legal protection by design~\cite{Hildebrandt:2017}, which should both preserve and strengthen contestabilty, i.e.\ the ability to contest not only the output of the data processing pipeline but also the very explanations produced to account for the functioning of the system~\cite{Hildebrandt:2023}.

\begin{figure}[tb]
	\centering
	\begin{tikzpicture}
		\node[anchor=south west,inner sep=0] (image) at (0,0) {\includegraphics[width=0.7\textwidth]{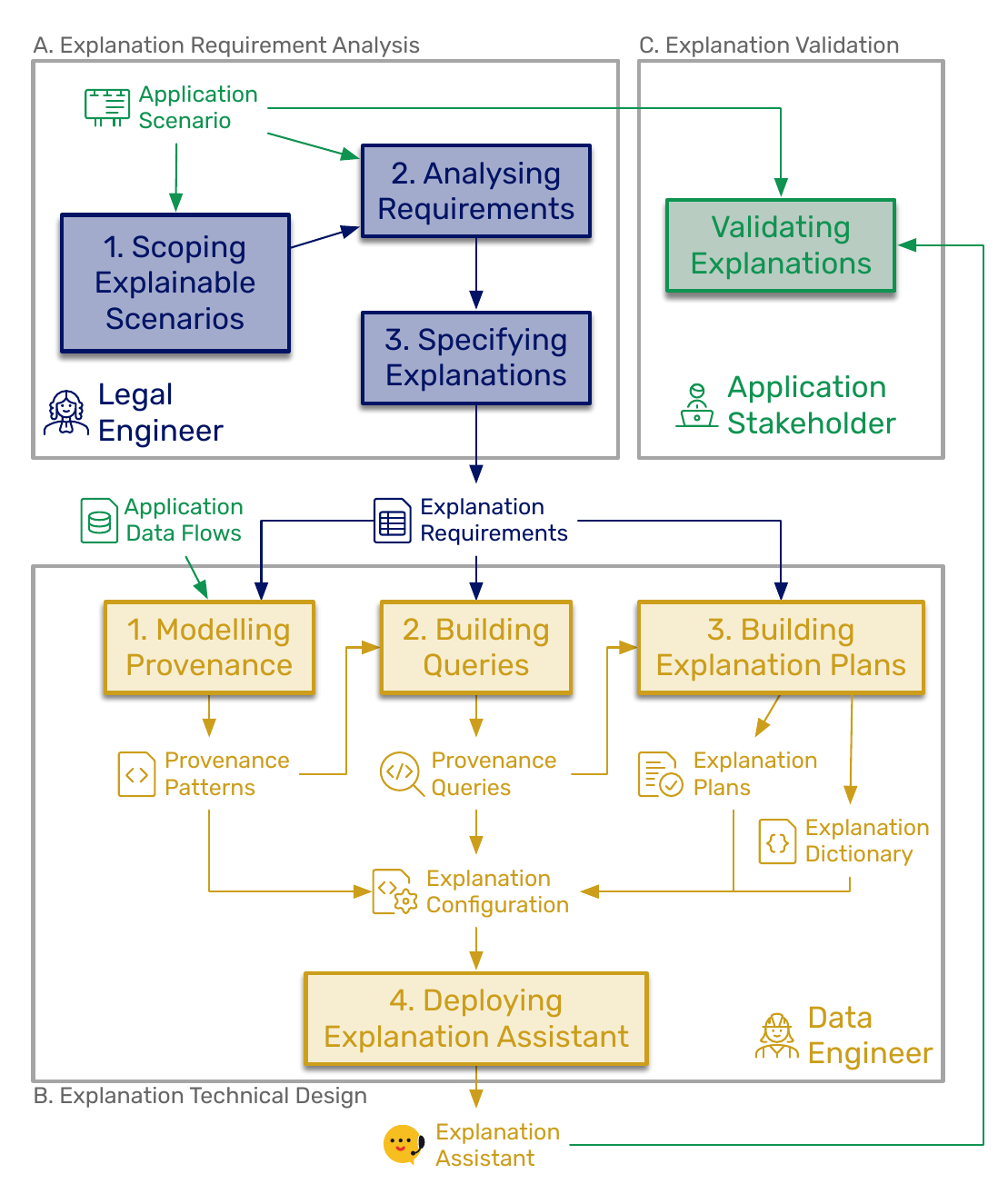}};
		\begin{scope}[x={(image.south east)},y={(image.north west)}]
			\draw[red,ultra thick,rounded corners] (0.33,0.75) rectangle (0.6,0.92) node[pos=0.15,yshift=-0.7ex] {\tiny A.2};
		\end{scope}
	\end{tikzpicture}
	\caption{Example of an Explainability-by-Design methodology}%
	\label{fig:pleadmethodology}
\end{figure}

An explainability-by-design methodology should follow a socio-technical process whereby explanation requirements are used to produce explanation statements that are then tested and validated with stakeholders before being translated into system requirements. This led us to revisit socio-technical integration research practices, which consist in embedding social scientists into engineering team. We thus decided to add a new role, that of the legal engineer~\cite{Freeland:2021}, whose function is to bridge between the compliance team and the engineering team. Indeed, not only is embedding social scientists within engineering teams important to identify social values midstream while engineers make decisions (e.g., transparency, understandability), but translating high-level legal and compliance requirements into fine-grained explanation requirements before generating system requirements is a fundamental intermediate step as well.  

As alluded above, explanations can have a variety of goals, including justification and transparency. In this paper, we conceive an explanation as a statement that provides the details or reasons relating to an output or decision. A decision is thus defined as any output of a data processing pipeline, produced after consideration of one or more data points. The output or decision can be followed by an action to bring about a correction or a resolution to a situation.  

It is possible to distinguish three phases within an explainability-by-design methodology~\cite{Tsakalakis_2022}, as illustrated in \cref{fig:pleadmethodology}:
\begin{enumerate}
    \item An \textit{Explanation Requirement Analysis} phase, depicted inside rectangle A, where legal engineers determine the explanation requirements of the processing in question; 
    \item An \textit{Explanation Technical Design} phase, depicted inside rectangle B, consisting of a set of technological steps, initiated by data engineers, to instantiate a service, the `Explanation Assistant', that generates explanations according to phase 1; and
    \item A \textit{Validation} phase, depicted inside rectangle C, where the generated explanations are evaluated by the stakeholders for their suitability and effectiveness.
\end{enumerate}

This paper focuses on the \textit{Explanation Requirement Analysis} phase depicted in \cref{fig:pleadmethodology} and specifically on step A2 of \cref{fig:pleadmethodology}  `\textit{Analysing Requirements.}'
It broadens the approach to explanaibility by focusing upon automated decision-making pipelines instead of being purely model-centric~\cite{Sokol2021} and is therefore of relevance in cases in which decision-making pipelines are only partially automated, like in the school allocation use case described below. The paper thus presents a typology developed to enable practitioners to systematically generate explanations of decisions through the formulation of explanation requirements, with the aim of putting the explanation recipient at the centre of the design in a `reflective sociotechnical approach'~\cite{Ehsan2020}. 
The typology does not represent an exhaustive set of categories and, therefore, remains open. It has been built out of a minimum set of distinct dimensions which reflect different concerns, e.g., the timing, form, content, and effect of explanations, that have been evaluated in the context of small-scale pilot studies. These dimensions are non-orthogonal in the sense that the selection of a particular value within one dimension may determine the selection of a particular value within another dimension. This typology is used to produce explanation requirements that are then mapped  to relevant data points within an audit trail so that it can be consumed by an explanation assistant. The typology is thus associated with categorisation rules that take as input applicable explanation requirements and output explanation specifications. 
The typology can be represented in machine-readable formats, such as a light ontology, which is included in the appendix of this paper.

The paper is organised into five sections. Following the introduction in \cref{section:introduction},
\cref{section:methodology} presents the methodology that was followed to develop the typology of explanations.
\Cref{section:typology} describes each dimension of the typology and associated criteria.
\Cref{section:discussion} illustrates its application in the context of small-scale pilot studies comprising two use cases.
\Cref{section:evaluation} discusses evaluation and \Cref{section:limitation} the limitations of the methodology and typology.
Finally, \cref{section:conclusion} highlights the potential of the typology as a classifier of explanations and discusses its limitations.

\section{Methodology}\label{section:methodology}

The explanation typology is the product of desk research and co-creation research techniques implemented to generate requirements from interactions with domain experts representing industry partners. 
The explanation typology is the work-product of a legal engineer who has been embedded within the engineering team after having worked closely with compliance teams of industry partners to translate high-level legal and compliance requirements into fine-grained explanation requirements.

Overall, the typology was built using 
the method by \citet{Nickerson:2013}.
First, the goals of the typology were set,
as depicted in \cref{tab:methodology} `1. Meta-Characteristics'.
The goal of the typology is to map the information needed to meet explanation requirements derived from an analysis of applicable legal or regulatory obligations, compliance rules or business policies.

\begin{table}[tb]
	\centering
	\resizebox{\textwidth}{!}{%
		\begin{tabular}{@{}>{\raggedright}p{.16\textwidth}p{\textwidth}@{}}
			\toprule
			\begin{tabular}{@{}l}
				\textbf{Methodology}\\ \textbf{Components}
			\end{tabular} & \textbf{Application} \\ \toprule
			\vspace{-5em}1. Meta-Characteristics & \parbox{\textwidth}{\begin{enumerate}[leftmargin=0cm,itemindent=0cm,labelwidth=\itemindent,labelsep=0cm,align=left,wide=0em, label={\tiny}, noitemsep,partopsep=0pt,topsep=0pt,parsep=0pt, font=\normalsize]
					\item []\textbf{Bases:} Information to meet the explanation requirements arising out of applicable regulatory obligations, business needs or other policies 
					\item []\textbf{Expected User:} Organisations determining their processing methods / System vendors designing processing systems for third-parties 
					\item []\textbf{Expected Use:} The proactive production of explanations as part of a supportive automated compliance strategy and the design of 'Explainable by Design' systems
					\item \textbf{Purpose:} The categorisation of information to produce comprehensive explanations for the applicable explanation requirements of the Organisation \end{enumerate}}\\ \midrule
			\vspace{-4.2em}2. Ending Conditions & \parbox{\textwidth}{\textbf{Objective Conditions:}
				\begin{enumerate}[wide=1.5em, leftmargin=3.5em, label=\footnotesize2.\arabic*., itemsep=0pt, parsep=0pt, font=\normalsize]
					\item A representative sample of objects has been examined
					\item No new dimensions or characteristics were added in the last iteration
					\item No dimensions or characteristics were merged or split in the last iteration
					\item Every dimension is unique and not repeated (i.e., there is no dimension duplication)
					\item Every characteristic is unique within its dimension (i.e., there is no characteristic duplication within a dimension)
			\end{enumerate}} \\
			& \textbf{Subjective Conditions:} Conciseness, Robustness, Comprehensiveness, Extensibility, and Explainability \\ \midrule
			\vspace{-2.5em}3. Empirical to Conceptual Approach &  \parbox{\textwidth}{\begin{enumerate}[wide=0em, leftmargin=*, itemsep=0pt, parsep=0pt, font=\normalsize, label=\footnotesize3.\arabic*.] 
					\item The sampling of two practical use cases of (semi-)automated decision making (loan application and school allocations) and their applicable rules
					\item The analysis and coding of `bases' from the applicable rules using desk research
					\item The categorisation of the `bases' into characteristics and dimensions\end{enumerate} }\\ \midrule
			\vspace{-2em}4. Conceptual to Empirical Approach & \parbox{\textwidth}{\begin{enumerate}[topsep=0pt, wide=0em, leftmargin=*, itemsep=0pt, parsep=0pt, font=\normalsize,label=\footnotesize4.\arabic*.]  
					\item The identification of explainability characteristics in the literature
					\item Comparing the explainability characteristics to own analysis
					\item The reorganisation of the typology taking into account the findings of step 4.2.\end{enumerate} }\\ \bottomrule
		\end{tabular}%
	}
	\caption{The \citet{Nickerson:2013}'s methodology as applied.}
	\label{tab:methodology}
\end{table}

Following an empirical-to-conceptual approach, an experimentation was performed in the context of two use cases. 
Pilot studies with two groups of stakeholders participating to two different use cases were conducted to generate a list of relevant explanation requirements in these sectors. 
The first use case concerned an automated credit scoring process leveraged to assess loan applications in the financial sector. 
The second use case involved a semi-automated process for allocating school places to UK pupils during annual school admission rounds. 
Each use case involved producers and consumers of explanations, i.e., decision makers from institutions that deploy automated decision pipelines and who are under an obligation to explain the output to end-users. 

The pilot studies were conducted as a proof of concept to demonstrate the feasibility and usefulness of categorising explanation requirements and transforming the output of the process into a machine-readable classifier.
Domain experts representing industry partners participated to the construction of explanation requirements together with the research team as well as to the production of explanation outputs, i.e.\ explanations.
Interviews were run with domain experts and outputs were discussed in the context of focus groups. Interview responses were analysed to reveal common themes.
The themes, or `bases', of the explanation requirements were grouped under different headings which were generalised to form dimensions of concern.
Subsequently, in a conceptual-to-empirical approach, the categories and dimensions produced in the experimentation stage were compared to explanation properties identified in the literature.
The typology was refined to ensure alignment with desk research findings and to make it more concise. 

The pilot studies confirmed the feasibility and usefulness of computing explanations on the basis of the typology, which was leveraged to produce a light ontology and an explanation assistant. Given the limited scope of the pilot studies, we do not claim that the typology amounts to an exhaustive taxonomy. We however assessed both the methodology and the typology in the light of the newly adopted EU AI Act: we found alignment between the methodology, the typology and the data governance requirements for high-risk AI systems\footnote{AI Act, Article 10} and the newly-introduced right to explanation.\negmedspace\footnote{AI Act, Article 86}

Several iterations were performed, during which the research team progressively added new dimensions until both objective and subjective ending conditions were finally met for the particular use cases.
The objective conditions were fulfilled when new iterations no longer split or merged dimensions or categories; when each dimension was unique and did not repeat; and, when each category was unique within a specific dimension. The subjective criteria were met by the last iteration.
The typology dimensions were within the \(7 \pm 2\) limit, set by \citet{Miller:1956}, which achieved conciseness.
Each category within each dimension was reduced to an elementary level to ensure differentiation of the object under investigation. The comprehensiveness (or relative completeness) of the typology, which implies that all dimensions are included and that new objects can be categorised, was tested through the production of examples based on explanation requirements derived from the GDPR and the School Admissions Code. The typology has been designed to be extensible and cover requirements stemming from a variety of frameworks. Finally, the explainability criterion, which evaluates how well the dimensions explain the object of the typology, was addressed in the context of focus groups with domain experts.

\section{Building a typology for explanations}\label{section:typology}

The overall goal of the explanation typology is to support a categorisation process that is able to produce a comprehensive set of explanation requirements from which it is then possible to identify building blocks for the computation stage. 

The sources of explanation requirements can vary. They can either be mandated by law \req{1}; or be grounded upon mere internal requirements, such as company policies \req{2}. Co-creation partners stressed the necessity to deploy explanations in different contexts \req{3}, to be able to scale and reuse explanation outputs easily \req{4}, and to be able to create them proactively before receiving requests from data subjects or addressees \req{5}. Issuers and recipients should also be able to retrieve explanations quickly \req{6}, explanations should empower \req{7} and educate their recipients \req{8}, and ultimately increase the level of engagement with the service \req{9}. Some partners viewed explanations as a tool to automate their transparency obligations \req{10}, even though they had to strike a trade-off between transparency and the protection of intellectual property including trade secrets \req{11}. Finally, explanations should be fit for purpose and to the point so as not to overwhelm their audience \req{12}.

All these considerations formed our first set of requirements. These requirements were then conflated and codified into `bases' to form typology dimensions. Each dimension describes a facet of an explanation. When needed, dimensions are broken down into sub-categories to provide sufficient granularity to the description. Each dimension is hierarchical. The depth of each dimension can vary. 

\Cref{dimensions} shows the nine dimensions.
\begin{figure}[tb]
	\centering
	\adjustbox{max width=.6\textwidth}{
		\begin{tikzpicture}[node distance = 5cm, auto]
			\tikzset{sun/.style={circle,
					color = black!80,
					fill = black!15,
					font=\large,
					draw=black!70,line width=.2pt,
					drop shadow,
					minimum size = 3.2cm,
					inner sep = 0.1cm}}
			\tikzset{satellite/.style={ellipse,
					color = black!80,
					draw=black!50,line width=.2pt,
					drop shadow,
					text width=2.2cm,
					align=center,
					minimum size = 1cm,
					inner sep = 0.1cm}}
			\tikzset{satellitearrow/.style={-latex, 
					line width = 0.11cm}}
			\pgfkeys{/pgf/regular polygon sides=9,/pgf/minimum size=0.75cm}
			\pgfnode{regular polygon}{center}{}{}{}
			\pgfusepath{draw}
			\draw (0,0) node (frame) [shape=regular polygon, minimum size=7cm, rotate=0] {};
			\node (sun)         at (frame.center)   [sun,fill=gray!10]                                           {Explanations};
			\node (satellitea)  at (frame.corner 1) [satellite, fill=red!10,xshift=0cm]                     {Source};
			\node (satelliteb)  at (frame.corner 2) [satellite, fill=cyan!10,xshift=-1cm]  {Criticality};
			\node (satellitec)  at (frame.corner 3) [satellite, fill=blue!10, xshift=-1cm]                      {Intended Recipient};
			\node (satellited)  at (frame.corner 4) [satellite, fill=green!10, xshift=-1cm]   {Explainability goal};
			\node (satellitee)  at (frame.corner 5) [satellite, fill=orange!10, xshift=.0cm]                     {Scope};
			\node (satellitef)  at (frame.corner 6) [satellite, fill=yellow!10,text width=2cm, xshift=1.3cm,yshift=.2cm]                     {Content};
			\node (satelliteg)  at (frame.corner 7) [satellite, fill=cyan!10,text width=2cm, xshift=1.1cm,yshift=.3cm]                     {Trigger};
			\node (satelliteh)  at (frame.corner 8) [satellite, fill=blue!10,text width=2cm, xshift=1cm]                     {Autonomy};
			\node (satellitei)  at (frame.corner 9) [satellite, fill=orange!10,text width=2cm, xshift=1cm]                     {Timing};
			
			\draw [satellitearrow, draw=red!20]  (sun) -- (satellitea);
			\draw [satellitearrow, draw=cyan!20]    (sun) -- (satelliteb);
			\draw [satellitearrow, draw=blue!20]   (sun) -- (satellitec);
			\draw [satellitearrow, draw=green!20]     (sun) -- (satellited);
			\draw [satellitearrow, draw=orange!20]  (sun) -- (satellitee);
			\draw [satellitearrow, draw=yellow!30]  (sun) -- (satellitef);
			\draw [satellitearrow, draw=cyan!20]  (sun) -- (satelliteg);
			\draw [satellitearrow, draw=blue!20]  (sun) -- (satelliteh);
			\draw [satellitearrow, draw=orange!20]  (sun) -- (satellitei);
		\end{tikzpicture}
	}
	\caption{Dimensions of explanations.}\label{dimensions}
\end{figure}

\begin{enumerate}[partopsep=.5em,]
	\item []\textbf{\dimension{Source}} defines the origin of the explanation requirement \req{1}.
	\item []\textbf{\dimension{Timing}} describes the moment at which the explanation is generated in relation to its object \req{10}.
	\item []\textbf{\dimension{Autonomy}} distinguishes between explanations that are generated without input and explanations that require an input for their generation \req{5}.
	\item []\textbf{\dimension{Trigger}} defines the events that trigger the generation of an explanation \req{2,5}.
	\item []\textbf{\dimension{Content}} captures the details about the formulation of an explanation \req{6,9,11}.
	\item []\textbf{\dimension{Scope}} assesses whether an explanation addresses a limited set of circumstances or has extended applicability \req{4}.
	\item []\textbf{\dimension{Explainability goal}} defines the functioning purpose of the generated explanation \req{3,7,8}. 
	\item []\textbf{\dimension{Intended Recipient}} defines the target audiences of the generated explanations \req{3}.
	\item []\textbf{\dimension{Criticality}} specifies whether the explanation is mandatory or recommended \req{2}.
\end{enumerate}

\subsection{Source}
\begin{figure}[b]
	\centering
	\begin{forest}                      
		typology, for root={
			ellipse,
			draw,
			parent anchor=east,},
		[Source
		[Tertiary,top color=col1,bottom color=col1,]
		[Secondary,top color=col1,bottom color=col1,]
		[Primary,
		[Implicit,,l=10pt, fork sep=15pt]
		[Explicit]]]
	\end{forest}
	\caption{The \dimension{Source} dimension}\label{type}
\end{figure}

The \dimension{Source} defines 
the origin of the explanation requirement (\cref{type}).
An explanation requirement may stem from three types of sources:
\begin{enumerate*}[label=(\roman*)]
	\item applicable laws;
	\item related authoritative guidance and standards; and, 
	\item internal compliance or business needs. 
\end{enumerate*}

When an explanation requirement is derived from the law, be it a piece of legislation or case law (i.e.\ primary sources), they are considered to be primary requirements. Such obligations can arise from a multitude of areas, e.g.\ public administrative law, consumer law, data protection law. Non-compliance with explicit explanation requirements can result in severe consequences for the organisation, such as reputational damage, fines and litigation costs. Therefore, organisations should have strong incentives to achieve compliance with explicit primary explanation requirements. In other circumstances, explanation requirements are merely implicit. This may be the case when compliance is de facto facilitated by the generation of an explanation. In other words, the generation of an explanation is characterised as good practice for achieving compliance.

\begin{example}{Primary explanation requirements}
	\textbf{Explicit:} School admission authorities in England must explain the reasons behind the refusal of a school place to a child when informing the parent (applicant) of the decision, under paragraph 2.32 of~\cite{DepartmentEducation:2021}.\\
	\textbf{Implicit:} Data controllers, who must be able to demonstrate compliance with the GDPR under Article 5(2), are under an implicit regulatory requirement to detect whether the processing is performed according to internal policies, which may be achieved through the generation of explanations. 
\end{example}

Because primary requirements for explanations come from legal rules, which are often high-level and open-ended, the entities that are called to comply with explanation requirements are often left with a margin of appreciation to determine how best to comply. The translation of high-level rules (e.g., ``data subjects should be informed about the logic involved in the processing of their personal data'') into secondary requirements (e.g., ``data subjects should be informed about whether a particular data point has contributed to the automated decision'') imply that a decision has been made to interpret the high-level requirement into context. Such translation can be facilitated by the release of guidance and standards, such as by national supervisory authorities and (supra-)national standard bodies. 

\begin{example}{Secondary explanation requirements}
	The Information Commissioner's Office, the national supervisory authority for data protection in the UK, has released authoritative guidance on explaining AI decisions in practice in collaboration with the Alan Turing Institute. The \citet{ICO:2021a} provides different tasks that organisations need to consider when deciding how to explain their AI systems. 
\end{example}

In practice, organisations tend to establish internal compliance functions. The complexity of these functions usually depends on the sector in which the organisation operates and its size. Internal compliance policies consist in translating primary and secondary requirements into actionable and auditable tertiary requirements. Such tertiary requirements are usually more granular than primary and secondary requirements.

\begin{example}{Tertiary explanation requirements}
	Internal requirements for explanations may arise out of designated roles and responsibilities (e.g.\ responsible agents), internal guidelines on the legalities of explanations created by in-house legal teams, standardised processes (e.g.\ quality control, email templates for correspondence with recipients), monitoring mechanisms and delivery systems for explanations (e.g.\ technical infrastructure). 
\end{example}

\subsection{Timing}
\begin{figure}[b]
	\captionsetup{singlelinecheck=false}
	\centering
	\begin{minipage}[b]{.5\textwidth}
		\centering 
		\begin{adjustbox}{max width=\textwidth}
			\begin{forest}                      
				typology, for root={
					ellipse,
					draw,
					parent anchor=east,},
				where level=1{top color=col1,bottom color=col1,	draw=black}{},
				[Timing
				[Ex post]
				[Ex ante]
				]
			\end{forest}
		\end{adjustbox} 
		\captionof{figure}{The \dimension{Timing} dimension}\label{time}
	\end{minipage}\hfill
	\begin{minipage}[b]{.5\textwidth}
		\centering 
		\begin{adjustbox}{max width=\textwidth}
			\begin{forest}                      
				typology, for root={
					ellipse,
					draw,
					parent anchor=east,},
				where level=1{top color=col1,bottom color=col1,	draw=black}{},
				[Autonomy
				[Reactive,name=C]
				[Proactive, name=A]
				]
			\end{forest}
		\end{adjustbox} 
		\captionof{figure}{The \dimension{Autonomy} dimension}\label{autonomy}
	\end{minipage}\hfill
\end{figure}

The \dimension{Timing} dimension expresses the moment at which an explanation is generated in relation to the object explained by the explanation (\cref{time}). 
An explanation can be generated either \textit{ex ante}, i.e. before the object(s) explained within the explanation take(s) place, 
or \textit{ex post}, i.e. after the object(s) materialise(s).

\begin{example}{Ex ante \& ex post explanations}
	For example, an explanation about the details of 
	the processing of personal data will be triggered \textit{\textbf{ex ante}}, 
	before the processing takes place, and will explain the intended processing that \textit{will} take place.
	On the contrary, explanations about a decision will be triggered \textit{\textbf{ex post}},
	after the decision has been reached and will explain the decision.
\end{example}

\subsection{Autonomy}
The \dimension{Autonomy} dimension distinguishes between explanations that are generated without any input from the recipient or 
explanations that are generated only as a response to specific input from the recipient. It is depicted in \cref{autonomy}.
A \dimension{Reactive} explanation is responsive.
Reactive explanations will typically be triggered by a direct request from an individual
or organisation for information about a process, event or decision.

\begin{example} {Reactive explanations}
	{An individual receives an unexpected or undesired decision (e.g.\ a lower credit score than expected) and, therefore, decides to contact customer support services for more information. At the point of contact (e.g.\ on receipt of the call or email), the customer support agent has to provide the reasons for this particular decision to the individual. The explanation is therefore triggered when the individual (subject to the decision) contacts the organisation (who made the decision) to obtain an explanation.}
\end{example}

On the other hand, \dimension{Proactive} explanations anticipate the need of the intended recipient for an explanation.
Proactive explanations will be triggered as soon as the processing, event, or (a part of the) decision-making process is fulfilled.

\begin{example}{Proactive explanations}
	An individual receives an unexpected or undesired decision (e.g.\ a lower credit score than expected). The decision is accompanied by an explanation, providing the reasons this decision has been taken, the data sources that were taken under consideration during the decision-making process, a brief overview of the decision-making process and the possible effects the decision might have on the individual. 
\end{example}

\subsection{Trigger}

The \dimension{Trigger} dimension (\cref{trigger}) expresses the event(s) that trigger(s) the generation of an explanation. As regards reactive explanations, the triggering event will typically be an action performed by the explanation recipient. However, explanations can also be triggered in the absence of a request or input from the recipient of the explanation, like in the example of an automated decision mentioned above. Triggering events comprise data processing activities such as the collection of an input to a decision-making pipeline, or the generation of an output from a decision-making pipeline.

\begin{example}{Triggering events}
	\textbf{Action:} a request by a customer to switch telephone providers, triggering an explanation about the necessary time to fulfil the request.\\
	\textbf{Processing:} the processing of personal data, triggering the production of explanations to the data subject(s) about the details of processing (even before a decision is reached).\\
	\textbf{Decision:}	the rejection of a loan application, triggering the production of an explanation about the factors that disqualified the applicant.
\end{example}

\subsection{Content}
\begin{figure}[b]
	\captionsetup{singlelinecheck=false}
	\centering
	\begin{minipage}[b]{.5\textwidth}
		\centering 
		\begin{adjustbox}{max width=\textwidth}
			\begin{forest}                      
				typology, for root={
					ellipse,
					draw,
					parent anchor=east,},
				where level=1{top color=col1,bottom color=col1,	draw=black}{},
				[Trigger
				[Action, name=B]
				[Process, name=F]
				[Decision, name=D]]
			\end{forest}
		\end{adjustbox} 
		\captionof{figure}{The \dimension{Trigger} dimension}\label{trigger}
	\end{minipage}\hfill
	\begin{minipage}[b]{.5\textwidth}
		\centering 
		\begin{adjustbox}{max width=\textwidth}
			\begin{forest}                      
				typology, for root={
					ellipse,
					draw,
					parent anchor=east,}
				[Content
				[Confidentiality
				[Disclosed]
				[Restricted]]
				[Minimum Content,top color=col1,bottom color=col1,]
				[Sensitivity
				[Personal Data
                [Directly Identifying]
                [Non-Directly Identifying]
                ]
				[Generalised Values]]
				]
			\end{forest}  
		\end{adjustbox} 
		\captionof{figure}{The \dimension{Content} dimension}\label{min-con}
	\end{minipage}\hfill
\end{figure}

The \dimension{Content} dimension (\cref{min-con}) aims to capture details to be included within the formulation of an explanation. The \dimension{Sensitivity} sub-category defines the sensitivity level associated with the content of the explanation. When explanations contain personal data it is important to label it as such. Hence the sub-category of \dimension{Personal Data}. The content of these explanations usually falls within the scope of data protection legislation and, as a result, their use is restricted and made conditional. Personal data can however be subject to a variety of techniques to minimise the data and reduce re-identification risks, such as generalisation, pseudonymisation and anonymisation techniques. Within the category of \dimension{Personal Data} it is often useful from a risk perspective to distinguish between \dimension{Directly Identifying} and \dimension{Non-Directly Identifying} data points. The categories \dimension{Generalised Values} and \dimension{Personal Data} are not mutually exclusive.\negmedspace\footnote{In Bridges, the UK Court of Appeal confirmed that the test for determining whether data is personal is whether ``the information recorded by [the controller] individuated [the individual] from all others, that is to say it singled him out and distinguished him from all others'' (on the application of Edward BRIDGES) [2020] EWCA Civ 1058, para. 46. In IAB Europe, the CJEU confirms that user preferences amount to personal data when the data may ``by reasonable means, be associated with an identifier, such as, inter alia, the IP address of that user’s device.'' CJEU C-604/22 IAB Europe v Gegevensbeschermingsautoriteit 7 March 2024 ECLI:EU:C:2024:214, para. 51}

\begin{example}{Sensitivity}
	\textbf{Generalised Values:} An explanation of age-restricted access rights may refer to age brackets (e.g. < 18, 21+ etc.) of user groups instead of the ages of the users.\\
	\textbf{Personal Data:} An explanation about the personal data that has been processed after the exercise from a data subject of their right to access will contain all the data that has been associated with an identified or identifiable individual, even if the particular data point is not directly or indirectly identifying.
\end{example}

The \dimension{Confidentiality} category distinguishes between \dimension{Disclosed} and \dimension{Restricted} explanations. The rationale for this category is to indicate whether an explanation is or can be disclosed to the public or whether access control restrictions should be put in place, e.g.\ because they amount to trade secrets or classified information.
The \dimension{Disclosed} and \dimension{Restricted} categories can be further specified to add granularity.\negmedspace\footnote{For example, by breaking down access levels in
	an ISO:27001 inspired format to
	confidential for senior management access;
	restricted for access by most employees;
	internal for access by all employees;
	and public information to which everyone has access.
	See~\cite{ISO:2013} for more details about the levels.}
The category \dimension{Disclosed} is relevant for all explanations that can be accessed by external actors and Restricted should apply to explanations that are intended for internal use only. For both types of explanations however, access should be determined on the basis of the need-to-know and least privilege principles, unless disclosed explanations are meant to be publicly available. 

\begin{example}{Confidentiality}
	\textbf{Disclosed explanation:} An explanation about a decision to reject an application for a loan because of a low credit score.\\
	\textbf{Restricted explanation:} The weights used by a credit scoring model in the applicants' data to calculate the specific credit score that led to the rejection of the application.
\end{example}

The \dimension{Minimum Content} category contains the types of information that must be included within the explanation to comprehensively address the explanation requirement. The categories of this dimension are intimately related to the domain and effect of the explanation requirement. The \dimension{Minimum content} is thus usually requirement-specific and it is not possible to generalise further at this stage.
Explanations must contain information to cover all the necessary bases for the explanation requirement they intend to address. The Minimum Content category correlates with the \dimension{Source} dimension as well as the \dimension{Sensitivity} and \dimension{Confidentiality} dimensions. For example, an explanation containing intellectual property data classified as confidential might need to undergo further processing to obfuscate the data in question before being disclosed to third parties. The \dimension{Minimum Content} is also correlated with the \dimension{Intended Recipient}. \citet{Sokol2021} explains, for example, that ``a system designed as a certification tool will usually reveal more information than one providing explanations to customers.''

\begin{example}{Minimum content}
	An organisation making automated decisions without human intervention about individuals under Article 22 of the GDPR must provide an explanation about the decision to the subject of the decision. According to~\cite[pp. 18--19]{ICO:2018} the explanation shall contain at minimum:
	\begin{enumerate}[itemsep=0em]
		\item which information was taken into account;
		\item the rationale behind the decision;
		\item the key decision points that formed the basis for the decision;
		\item any alternative decisions that were considered and why they were not preferred;
		\item how to ask for a review;
		\item how to make an appeal and the available appeal grounds.
	\end{enumerate}
\end{example}

\subsection{Scope}

The \dimension{Scope} dimension (\cref{scope})
assesses whether the explanation only applies to a particular case or whether it can be reused in other contexts.
\dimension{Local} explanations are context-specific. For example, when the recipient changes, the content will have to change as well.
In contrast, a \dimension{Universal} explanation remains the same each time the decision-making pipeline produces an output. 

\begin{figure}[h]
	\centering
	\begin{adjustbox}{max width=1\textwidth}
		\begin{forest}                      
			typology, for root={
				ellipse,
				draw,
				parent anchor=east,},
			where level=1{top color=col1,bottom color=col1,	draw=black}{},
			[Scope
			[Local]
			[Universal]]
		\end{forest}  
	\end{adjustbox}
	\caption{The \dimension{Scope} dimension}\label{scope}
\end{figure}

\begin{example}{Scope}
	\textbf{Local:} The over-subscription criteria of each of the selected schools in a school admission application will result in an explanation whose scope will be limited to the application in question. \\
	\textbf{Universal:} The contact details of the Office of the School Adjudicator remain immutable. An explanation about how to contact the Office to challenge a decision will have applicability across all school allocation decisions. \\
\end{example}

\subsection{Explainability Goal}
\begin{figure}[b]
	\centering
	\begin{minipage}[t]{\textwidth}
		\centering
		\begin{forest}                      
			typology, for root={
				ellipse,
				draw},
			[Explainability goal
			[Understandability
			]
			[Intervenability]
			]
		\end{forest} 
		\caption{The \dimension{Explainability goal} dimension}\label{goal}
	\end{minipage}
\end{figure}

The \dimension{Explainability Goal} dimension (\cref{goal}) defines the functional purpose(s) that the generation of a particular explanation intends to achieve. 
Contrary to what is often assumed, explanations can serve a variety of purposes. The meaningfulness of an explanation, which is dependent upon its wording and format, is intimately related to the specific goal it pursues. To be both compliant and effective, an explanation must enable the recipient to understand a decision and/or a particular aspect of a decision-making process, and, when necessary, to take action. Explainability goals can thus be divided into two families: \dimension{Understandability} goals and \dimension{Intervenability} goals.
The two families reflect the main ``cognitive effects'' of explanations on recipients, outlined in~\cite{Maybury:1992} and~\cite{Chander:2018}.\negmedspace\footnote{\citet{Maybury:1992} identifies ``entity knowledge'', ``event and state knowledge'' and ``know-how knowledge'' relating to understandability and ``change beliefs/evoke action'' relating to intervenability, whereas \citet{Chander:2018} identify ``education'' and ``trust'' on the one hand and ``design'', ``action'' and ``trouble-shooting'' on the other.}
By way of example, the goals of justification and transparency are two sub-categories of understandability.

Goals in the \dimension{Understandability} family (\cref{understandability}) aim 
at fostering either trust or confidence in the decision output or the decision-making pipeline. Understandability is key in building trust for systems where decision-making can
be highly consequential (e.g.\ because it produces legal effects), and
the correctness or validity of the decisions
cannot be readily verified~\cite{Ye1995}.
We have identified eight of those goals:

\begin{enumerate}[label={}]
  \item \textbf{Performance},  assessing the performance of the system;	
  \item \textbf{Accountability}, explaining whether and how the system owner is in control; 
	\item \textbf{Accuracy}, reporting on the accuracy of the input data; 
	\item \textbf{Consequences}, explaining the processing impact upon the recipient;
	\item \textbf{Data minimisation}, explaining how the input data has been tailored to the processing purpose;
	\item \textbf{Fairness}, explaining how biases and discrimination have been mitigated; 
	\item \textbf{Overall Transparency}, explaining the main steps of the processing system;
	\item \textbf{Justification}, explaining the reasons for the output.
\end{enumerate}

\begin{figure}[tb]
	\captionsetup{singlelinecheck=false}
	\centering
	\begin{minipage}[b]{.49\textwidth}
		\centering 
		\begin{adjustbox}{max width=\textwidth}
			\begin{forest}                      
				typology, for root={
					draw,
					parent anchor=east,},
				where level=0{top color=col2,bottom color=col2,	draw=black}{},
				where level=1{top color=col1,bottom color=col1,	draw=black}{},
				where level=2{top color=col1,bottom color=col1,	draw=black}{}, 	
				[Understandability
				[Justification]
				[Overall Transparency]
				[Fairness]
				[Data minimisation]
				[Consequences]
				[Accuracy]
				[Accountability]
                [Performance]
				]
			\end{forest}
		\end{adjustbox} 
		\captionof{figure}{The \dimension{Understandability} goals}\label{understandability}
	\end{minipage}\hfill
	\begin{minipage}[b]{.49\textwidth}
		\centering 
		\begin{adjustbox}{max width=1\textwidth}
			\begin{forest}                      
				typology, for root={
					draw,
					parent anchor=east,},
				where level=0{top color=col2,bottom color=col2,	draw=black}{},
				where level=1{top color=col1,bottom color=col1,	draw=black}{},
				where level=2{top color=col1,bottom color=col1,	draw=black}{},
				[Intervenability
				[Scrutability]
				[Restriction]
				[Additional information]
				[Data rectification]
				[Data portability]
				[Human intervention]
				[Behaviour modification]
				[Data erasure]
				[Contestability]
				[Data access]
				] 
			\end{forest}  
		\end{adjustbox}
		\captionof{figure}{The \dimension{Intervenability} goals}\label{intervenability}
	\end{minipage}
\end{figure}

Goals belonging to the \dimension{Intervenability} family (\cref{intervenability})  aim at empowering recipients to take action,\negmedspace\footnote{The concept of intervenability is borrowed from the literature on data protection goals~\cite{SDM3}.} e.g.\ to exercise corrective individual rights or more simply to monitor the performance of the system (e.g.\ trouble-shooting). The \dimension{Intervenability} family comprises ten goals:
\begin{enumerate}[label={}]
	\item \textbf{Data access}, accessing input data; 
	\item \textbf{Contestability}, challenging an output;
	\item \textbf{Data erasure}, deleting input and/or output data;
	\item \textbf{Behaviour modification}, modifying the behavior of the system;
	\item \textbf{Human intervention}, performing human review;
	\item \textbf{Data portability}, porting data to a third party;
	\item \textbf{Data rectification}, rectifying input/output data;
	\item \textbf{Additional information}, accessing additional information about the system;
	\item \textbf{Restriction}, restricting the use of the data or system;
	\item \textbf{Scrutability}, flagging errors.
\end{enumerate}

\dimension{Explainability goal} aims to highlight the purpose of the generated explanation 
so that its effectiveness and compliance can be assessed against the explainability requirements.
An explanation may serve multiple goals at once.
For example, an explanation about the use of automated decision-making 
aims at informing the recipient about the operation of the system (the \dimension{Transparency} goal)
but also holding the decision-maker accountable (the \dimension{Accountability} goal).
Explanations from different goals can also be combined together
into larger explanatory statements:
The explanation under Article 22 of the GDPR must explain the reasons and impact of an automated decision
(\dimension{Fairness}, \dimension{Overall Transparency} and \dimension{Consequences}),
but should also provide instructions on how to request a human review of the decision and appeal it
(\dimension{Human intervention} and \dimension{Contestability}).

\subsection{Intended Recipient}
\begin{figure}[htb]
	\begin{forest}                      
		typology, for root={
			ellipse,
			draw,
			parent anchor=east,} 
		[Intended Recipient
		[Inward-facing
		[Legal engineer]
		[Data engineer]
		[Business analyst]
		[Administrator]
        [Auditor]
		]
		[Outward-facing
		[Third party]
		[Supervisory authority]
		[Data subject]
        [Decision addressee]
		]
		]
	\end{forest}
	\caption{The \dimension{Intended Recipient} dimension}\label{recipient}
\end{figure}

The \dimension{Intended Recipient} dimension defines the  categories of recipients that will be associated with the explanations (\cref{recipient}).
Identifying the target audience of an explanation is essential to be able to assess understandability, as comprehension skills vary across groups of recipients. Besides, as \citet{Bohlender2019} argue, ``the notion of explanation is not absolute but rather relative to a target group''. 
This dimension draws a distinction between explanations targeting the public and explanations for internal consumption.
\dimension{Outward-facing} explanations can target the public, including data subjects and other decision addresses, supervisory authorities responsible for overseeing the organisation’s processing or other third parties that interface with the organisation. Explanations can also be \dimension{Internal-facing}, aiming to assist the employees of an organisation in monitoring and troubleshooting the processes and their compliance with applicable rules. At first glance, five broad categories of internal-facing recipients can be distinguished: administrators of the decision-making system; business analysts making decisions; legal engineers who determine the primary and secondary requirements and contribute to the elicitation of tertiary requirements; data engineers who are responsible for managing the data; and auditors, who are in charge of monitoring compliance. However, this dimension is highly dependent on the nature of the implementing organisation. An organisation could choose to produce explanations for third parties participating to the decision-making pipeline.

\begin{example}{Intended Recipients}
	\textbf{Outward-facing:} An organisation will need to provide explanations to subjects of automated decisions expressed in layman terms, and may want to disclose explanations to the supervisory authority to demonstrate how its operations comply with the law.\\
	\textbf{Inward-facing:} Explanations monitoring the automated operations of the organisation can address software engineers, in-house lawyers or business managers. Since these groups have different types of specialist knowledge, they will each require a type of explanation that takes into account the degree of their related expertise.
\end{example}

\subsection{Criticality}
\begin{figure}[t]
	\captionsetup{singlelinecheck=false}
	\centering
	\begin{adjustbox}{max width=\textwidth}
		\begin{forest}                      
			typology, for root={
				ellipse,
				draw,
				parent anchor=east,}, 
			where level=1{top color=col1,bottom color=col1,	draw=black}{},
			[Criticality
			[Mandatory]
			[Recommended]
			] 
		\end{forest}
	\end{adjustbox} 
	\caption{The \dimension{Criticality} dimension}\label{criticality}
\end{figure}

In practice, there are cases where an obligation to explain an action or omission arises on the basis of an applicable legal rule. However, there are cases where an organisation can decide that providing an explanation is beneficial even in the absence of mandatory requirement. The final dimension, \dimension{Criticality}, depicted in \cref{criticality}, captures the distinction between a mandatory and a recommended explanation requirement.

An explanation requirement is \dimension{Mandatory} where an applicable legal rule and/or a governance framework imposes an obligation to provide an explanation. On the contrary, an explanation is \dimension{Recommended} where it is not strictly speaking mandatory to generate an explanation to fulfil an obligation imposed by a legal and/or governance prescription. 

\begin{example}{Different criticality levels}
	\textbf{Mandatory:} Under the School Admissions Code, admission authorities who inform parents of a decision to refuse their child a place at a school are obliged to explain the reasons for the refusal.\\
	\textbf{Recommended:} A school admission authority may want to explain to parents that if they accept their second or third preference, their child will be placed on the waiting list for their first and eventually second preference.
\end{example}

\section{Producing explanations on the basis of the typology}\label{section:discussion}
A first round of explanations was produced by the research team and presented to the pilot study participants, who worked with the research team to fine tune them in an iterative process. Explanations were derived from both primary requirements, i.e., the GDPR, the School Admissions Code, and secondary requirements, i.e., relevant regulatory guidance.\negmedspace\footnote{Arising out of the Information Commissioner's guidance for the GDPR, for AI and for credit scoring.}

\subsection{Loan application}\label{subsection: loan application}

\begin{sidewaystable}
	\centering
	\resizebox{\textwidth}{!}{
		\begin{talltblr}[
			caption = {},
			label = {none},
			]{
				colspec = {@{}p{5cm}llllllllllllp{4cm}llllllll@{}}, width = \linewidth,
				row{odd} = {white}, row{even} = {red!45!white!15},
				row{1-3} = {green!50!black!15,font=\bfseries},
				hline{4,Z} = {1pt},
			}
			
			& \SetCell[r=1]{f}Source & \SetCell[c=2]{c}{\rotatebox{60}{Timing}} && \SetCell[c=2]{c}{\rotatebox{60}{Autonomy}} && \SetCell[c=3,r=1]{c,f}{Trigger} &&& \SetCell[c=5,r=1]{c,f}{Content} &&&&& \SetCell[c=2]{c}{\rotatebox{60}{Scope}} && \SetCell[c=2,r=1]{c,f}{Explainability goal} && \SetCell[c=2,r=1]{c,f}{Intended recipient} && \SetCell[c=2]{c}{\rotatebox{60}{Criticality}} \\ \cmidrule[lr]{3-4} \cmidrule[lr]{5-6} \cmidrule[lr]{7-9} \cmidrule[lr]{10-14} \cmidrule[lr]{15-16} \cmidrule[lr]{17-18} \cmidrule[lr]{19-20} \cmidrule[lr]{21-22} 
			&  & \rotatebox[origin=r]{90}{Ex ante} & \rotatebox[origin=r]{90}{Ex post} & \rotatebox[origin=r]{90}{Proactive} & \rotatebox[origin=r]{90}{Reactive} & \rotatebox[origin=r]{90}{Decision}  & \rotatebox[origin=r]{90}{Process} & \rotatebox[origin=r]{90}{Action} & \SetCell[c=2]{c}{\rotatebox[origin=r]{90}{Sensitivity}} && \SetCell[c=2]{c}{\rotatebox[origin=r]{90}{Confidentiality}} && Minimum Content & \rotatebox[origin=r]{90}{Universal} & \rotatebox[origin=r]{90}{Local} & Understandability & Intervenability & Outward-facing & Inward-facing & \rotatebox[origin=r]{90}{Mandatory} & \rotatebox[origin=r]{90}{Recommended} \\ \cmidrule[lr]{10-11}\cmidrule[lr]{12-13}
			&  &  &  &  &  &   &  &  & \rotatebox[origin=r]{90}{\parbox{2.1cm}{Non-Directly Identifying}} & \rotatebox[origin=r]{90}{\parbox{2.1cm}{Directly Identifying}} & \rotatebox[origin=r]{90}{\parbox{2.1cm}{Disclosed}} & \rotatebox[origin=r]{90}{\parbox{2.1cm}{Restricted}} & &  &  &  &  &  &  &  &  \\

			Existence of automated decision-making & Primary & $\bullet$ & $\bullet$  & $\bullet$ & $\bullet$  & $\bullet$   & $\bullet$ & $\bullet$  & $\bullet$ &  &$\bullet$& & Information about the logic \& significance of the decision & $\bullet$ &  & {Accountability \\ Overall Transparency \\Fairness} & Contestability  & Data subject &  & $\bullet$ &  \\
			
			Meaningful information about the logic & Primary & $\bullet$ & $\bullet$  & $\bullet$ & $\bullet$  & $\bullet$   & $\bullet$ & $\bullet$  & $\bullet$ &  &$\bullet$& & Information meaningful enough to facilitate the data subjects' exercise of their rights & $\bullet$ &  & {Accountability \\ Overall Transparency \\Fairness} & {Contestability \\ Scrutability}  & Data subject &  & $\bullet$ &  \\
			
			The categories of personal data used in the decision-making process & Primary & $\bullet$ & $\bullet$  & $\bullet$ & $\bullet$  & $\bullet$   & $\bullet$ & $\bullet$  & $\bullet$ &  &$\bullet$& & Categories (types) of data processed and their origin & $\bullet$ &  & {Accuracy \\Fairness} & {Rectification \\ Contestability} & Data subject &  & $\bullet$ & \\
			
			The relevance of the information for the decision-making & Secondary & $\bullet$ & $\bullet$  & $\bullet$ & $\bullet$  & $\bullet$   & $\bullet$ & $\bullet$  & $\bullet$ &  &$\bullet$& &  Relevance to the type of decision-making (e.g. transaction history as indication of expenditure) & $\bullet$ &  & {Data Minimisation \\Fairness} &  {Restriction \\ Deletion \\ Contestability} & Data subject &  & & $\bullet$\\

            Personal Data points  & Primary & $\bullet$ & $\bullet$  & $\bullet$ & $\bullet$  & $\bullet$   & $\bullet$ & $\bullet$  & & $\bullet$ &$\bullet$& &  Personal data values &  & $\bullet$ & {Accuracy \\ Fairness} & {Rectification \\ Contestability \\ Scrutability} & Data subject &  & $\bullet$  & \\
			
			The significance of envisaged consequences& Primary & $\bullet$ & $\bullet$  & $\bullet$ & $\bullet$  & $\bullet$   & $\bullet$ & $\bullet$  & $\bullet$ &  &$\bullet$& & Decision outcomes (e.g. offer or denial of loan) and associated consequences & $\bullet$ &  & {Consequences \\Fairness} & {Contestability} & Data subject &  & $\bullet$ &  \\
			
			Decision taken solely by automated means & Primary &  & $\bullet$ & & $\bullet$  & $\bullet$ &   &  & $\bullet$  & &$\bullet$& & Description of decision-making process &  $\bullet$ &  & {Accountability \\ Overall Transparency \\Fairness} & {Contestability} & Data subject &  & $\bullet$ &  \\
			
			Reasons for reaching the decision & {Primary:implicit \\ Secondary} &  & $\bullet$ & & $\bullet$ & $\bullet$  &  &  &  & $\bullet$ &$\bullet$& & Criteria used to reach the decision &  & $\bullet$ &  {Justification \\Fairness} & {Contestability} & Data subject &  & $\bullet$ &  \\
			
			Key data points  & {Primary:implicit \\ Secondary} &  & $\bullet$ & & $\bullet$ & $\bullet$ &   &  &  & $\bullet$ &$\bullet$& & Key data values &  & $\bullet$ & {Accuracy \\ Fairness} & {Rectification \\ Contestability \\ Scrutability} & Data subject &  &$\bullet$ &  \\

			Key associated weights  & Secondary &  & $\bullet$ & & $\bullet$ & $\bullet$ &   &  & $\bullet$  & &$\bullet$& & Associated weights & $\bullet$ &  & {Justification \\ Fairness} & {Contestability \\ Scrutability} & Data subject & &  &  $\bullet$\\

            Process to request a review of the decision & {Primary:implicit \\ Secondary}  &  & $\bullet$ & & $\bullet$ & $\bullet$ &   &  & $\bullet$  & &$\bullet$& & Necessary input data, contact info of reviewer & $\bullet$ &  & {Accountability \\ Consequence} & {Contestability \\ Human intervention} & Data subject & & $\bullet$& \\

            Response to the review request & {Primary:implicit \\ Secondary} &  & $\bullet$ &  & & $\bullet$  &  & $\bullet$ &  & $\bullet$ &$\bullet$&& Original facts; additional evidence; outcome of review &  & $\bullet$ & {Fairness \\ Overall Transparency \\ Accountability} & {Contestability \\ Human intervention} & Data subject &  & $\bullet$ & \\

		\end{talltblr}
	}
	\caption{Classification of explanation requirements related to GDPR Articles 13, 14, 15 and 22.}\label{tab:GDPR22-1}
\end{sidewaystable}

\begin{sidewaystable}
	\centering
	\ContinuedFloat
	\resizebox{\textwidth}{!}{
		\begin{talltblr}[
			caption = {},
			label = {none},
			]{
				colspec = {@{}p{5cm}llllllllllllp{4cm}llllllll@{}}, width = \linewidth,
				row{odd} = {white}, row{even} = {red!45!white!15},
				row{1-3} = {green!50!black!15,font=\bfseries},
				hline{4,Z} = {1pt},
			}
			
			& \SetCell[r=1]{f}Source & \SetCell[c=2]{c}{\rotatebox{60}{Timing}} && \SetCell[c=2]{c}{\rotatebox{60}{Autonomy}} && \SetCell[c=3,r=1]{c,f}{Trigger} &&& \SetCell[c=5,r=1]{c,f}{Content} &&&&& \SetCell[c=2]{c}{\rotatebox{60}{Scope}} && \SetCell[c=2,r=1]{c,f}{Explainability goal} && \SetCell[c=2,r=1]{c,f}{Intended Recipient} && \SetCell[c=2]{c}{\rotatebox{60}{Criticality}} \\ \cmidrule[lr]{3-4} \cmidrule[lr]{5-6} \cmidrule[lr]{7-9} \cmidrule[lr]{10-14} \cmidrule[lr]{15-16} \cmidrule[lr]{17-18} \cmidrule[lr]{19-20} \cmidrule[lr]{21-22} 
			&  & \rotatebox[origin=r]{90}{Ex ante} & \rotatebox[origin=r]{90}{Ex post} & \rotatebox[origin=r]{90}{Proactive} & \rotatebox[origin=r]{90}{Reactive} & \rotatebox[origin=r]{90}{Decision}  & \rotatebox[origin=r]{90}{Process} & \rotatebox[origin=r]{90}{Action} & \SetCell[c=2]{c}{\rotatebox[origin=r]{90}{Sensitivity}} && \SetCell[c=2]{c}{\rotatebox[origin=r]{90}{Confidentiality}} && Minimum Content & \rotatebox[origin=r]{90}{Universal} & \rotatebox[origin=r]{90}{Local} & Understandability & Intervenability & Outward-facing & Inward-facing & \rotatebox[origin=r]{90}{Mandatory} & \rotatebox[origin=r]{90}{Recommended} \\ \cmidrule[lr]{10-11}\cmidrule[lr]{12-13}
			&  &  &  &  &  &   &  &  & \rotatebox[origin=r]{90}{\parbox{2.1cm}{Non-Directly Identifying}} & \rotatebox[origin=r]{90}{\parbox{2.1cm}{Directly Identifying}} & \rotatebox[origin=r]{90}{\parbox{2.1cm}{Disclosed}} & \rotatebox[origin=r]{90}{\parbox{2.1cm}{Restricted}} & &  &  &  &  &  &  &  &  \\
			
			Key decision points  & Secondary &  & $\bullet$ & & $\bullet$ & $\bullet$ &   &  &  & $\bullet$ & & $\bullet$ & Key data values &  & $\bullet$ & {Accuracy \\ Performance \\ Accountability} & {Scrutability} & & {Administrator \\ Auditor} & &  $\bullet$\\
            
            Verification of the results & Secondary &  & $\bullet$ &  & $\bullet$v & $\bullet$  &  &  &  & $\bullet$ & &$\bullet$ & Processed data; data sources; model rules; audit trails &  & $\bullet$ & {Performance \\ Accountability} & Scrutability &  & {Administrator \\ Auditor} &  & $\bullet$ \\
			
			Business rules applied to the decision & Secondary &  & $\bullet$ & $\bullet$ &  & $\bullet$  &  &  &  $\bullet$ &  & & $\bullet$ & Business rules; processing purposes &  & $\bullet$ & {Accountability} & Scrutability &  & {Administrator \\ Auditor} &  & $\bullet$ \\
			
			Alternatives to the decision \& reasons for rejection & Secondary &  & $\bullet$ &  & $\bullet$ & $\bullet$  &  &  &  & $\bullet$ & & $\bullet$ & Alternative decisions; benefits of decision chosen or drawbacks of alternatives &  & $\bullet$ & {Performance \\ Accountability} & {Modifying a behaviour \\ Scrutability} &  & {Administrator \\ Business analyst \\ Auditor} &  & $\bullet$ \\

			Review carried out by qualified \& authorised reviewer & {Primary:impolicit \\ Secondary} &  & $\bullet$ &  & $\bullet$ &   &  & $\bullet$ &  & $\bullet$ & & $\bullet$& Contact info of reviewer; qualifications; track record & $\bullet$ &  & {Accountability} & Scrutability &  & {Administrator \\ Auditor} & $\bullet$ &  \\

		\end{talltblr}
	}
	\caption{Classification of explanation requirements related to GDPR Articles 13, 14, 15 and 22 (continued).}\label{tab:GDPR22-2}
\end{sidewaystable}

In this scenario, a prospective applicant, Alex, a UK resident, wishes to apply for a loan through the website of a bank located in the UK, the Bank.
The bank requests some information about the income and living habits of the applicant.
It then uses this information together with a credit score for Alex, received from a collaborating credit reference agency --- the CRA,
to calculate the creditworthiness of Alex. 
If Alex's creditworthiness is above a certain threshold, the application is automatically approved.
Below a certain threshold, the application is automatically rejected, whereas if it falls in between the thresholds, it is redirected to a manual assessment by an officer.
Because the Bank uses solely automated decision-making for applications that do not go to manual assessment,
UK GDPR Article 22 applies. 
 
UK GDPR Article 22 prohibits solely automated decision-making ingesting personal data of individuals if the decision produces legal or similarly significant effects for them. Exceptionally, solely automated decision-making can take place if
\begin{quote}
	\textit{the data controller [implements] suitable measures to safeguard the data subject’s rights and freedoms and legitimate interests, at least the right to obtain human intervention on the part of the controller, to express his or her point of view and to contest the decision.}
\end{quote}

Recital 71 specifies that suitable measures ``\textit{should include specific information to the data subject and the right }[\ldots] \textit{to obtain an explanation of the decision reached}''.
Although only persuasive at most, national courts within the EU have on occasions recognised such a right.\negmedspace\footnote{See~\cite{Uber:2021} at para. 4.39 ``In that case, the controller must still take appropriate measures, including at least the right to human intervention, the right of the data subject to make his point of view known and the right to challenge the decision (Article 22 (3) GDPR and recital 71 GDPR)''. See also~\cite{Daten:2020,person:2020} where the Austrian and Irish DPAs held that credit scoring companies were under an obligation to provide meaningful information about the calculation of the credit scores.}

Analysing UK GDPR Article 22, Recital 71 and other relevant provisions of the GDPR, we started by distinguishing five distinct explanation requirements. Organisations using solely automated decision-making to produce significant effects must:
\begin{enumerate}[label=\arabic*.]
	\item Provide information about the automated decision-making system to the subjects of the decisions;\label{GDPR-info}
	\item Provide an explanation about the decision reached;
	\item Allow the subject of the decision to express their point of view;
	\item Provide a way for the subject of the decision to obtain human intervention; and,
	\item Advise how they can challenge the decision. 
\end{enumerate}
 
Under UK GDPR transparency requirements, the information provided to the data subject, i.e. the future subject of the automated decision, 
must go beyond the mere existence of automated decision-making.
Data controllers must also supply ``\textit{meaningful information about the logic involved}'' as well as
the ``\textit{significance}'' and ``\textit{envisaged consequences}'' of any automated decision.\negmedspace\footnote{GDPR articles 13(2)(f), 14(2)(g) and 15(1)(h).}

The explanation mandate about the way the automated decision-making system works and its consequences comes on top of the obligation to provide information about the nature and purpose of the processing, the identity of the data controller and the available data subject rights.\negmedspace\footnote{See GDPR articles 13, 14 and 15.}
The UK GDPR does not give examples of what it means by ``\textit{meaningful}'',
but at the very least the information should be meaningful enough to facilitate the exercise of data subjects' rights \citep{Selbst:2017}.
In~\cite{ICO:2018}, the UK Information Commissioner
specifies additional requirements that are necessary to satisfy Article 22.
The meaningful information about the logic must also include the types of information used in making the decisions and why this information is relevant. In explaining a specific decision to a data subject, the organisation should explain how the decision was reached and its rationale. To assist organisations in generating comprehensive explanations, the Information Commissioner recommends that they keep records of the key decision points, the verification methods of the result, the business rules that apply to the decision and whether any alternatives existed. If there were alternatives, organisations should record why they were not preferred over the outcome of the decision.

An explanation is also required to enable the subject of an automated decision to exercise her rights. Article 22 requests clear communication of the right to ask for review, human intervention, express objections, and contest the decision. The Information Commissioner considers that it would make business sense to provide an explanation at the point of delivery of the decision to inform the subject about how to request a review or appeal of the decision. Organisations have to demonstrate that the review is carried out by a qualified reviewer who is authorised to assess the decision-making process. At the end of the review, they have to issue a formal response explaining the outcome of the review process.

Overall, we started by deriving GDPR-related explanation requirements. These requirements are not exhaustive and could be expanded by considering other articles or translating additional parts of the ICO's guidance. These have been classified in \cref{tab:GDPR22-1}.

The classification allows the identification of necessary components to construct a targeted explanation for each of the GDPR requirements. In addition, it assists in determining potential overlaps between requirements, or eventual complementarity.

Working with our partners, we decided to add explanations requirements for inward-facing monitoring and auditing, which are also recommended by the ICO. 
 
The remaining dimensions determine the delivery rules that dictate when and to whom the explanation will be presented.\negmedspace\footnote{%
	The details about the technical design and implementation are beyond the scope of this paper. A thorough explanation of the technical steps can be found in~\cite{Huynh:2021}.}
For the delivery of explanations, the chosen format follows the design guidelines identified by~\citet{Simkute:2021}.\negmedspace\footnote{See pp. 11--12, where the authors provide a list of design goals tuned to different expertise levels, e.g.\ the use of pop-up alerts; the use of structured information and visualisations; the ability to actively question the data; the provision of interactive manipulation; etc.}

We thus constructed explanations for the explanation requirements related to UK GDPR Articles 13, 14, 15 and 22, as illustrated in the coloured samples below, where the minimum content is highlighted. The samples are not set in stone and only aim at shedding light on the tension between data subjects' actual needs and controllers' own interests. 

To provide information about automated decision-making before the processing takes place (\textit{ex ante}) we produced the following explanations, based on the break down in \cref{tab:GDPR22-1}.

\NewTblrTheme{fancy}{
	\DefTblrTemplate{conthead}{default}{}
	\DefTblrTemplate{caption}{default}{}
	\DefTblrTemplate{capcont}{default}{(\textit{\ldots continued})}
	\DefTblrTemplate{contfoot-text}{default}{\textit{Continued on next page\ldots}}}

{\small
	\begin{longtblr}[
		theme=fancy,
		caption={},
		label={none},
		]{colspec={|@{}l @{}X@{}|}, rowspec={|QQ| |QQQ| |QQ| |QQ| |QQQ|},
			row{1,3,6,8,10}={bg=col2!90, font=\bfseries},}
		
		R1: & \hlcyan{Existence} of automated decision-making \\
		E1: & To be able to process applications quickly and accurately, the Bank \hlcyan{uses a solely automated system}.  \\ 
		
		R2: & Meaningful information \hlorange{about the logic}	\\
		E2.1: & The system \hlorange{will screen your application based on} the information you provide in the application form, information we hold on file about you and information we receive about you from our partner the CRA\@. \hlorange{Your information will be assessed against an affordability threshold, to determine whether you can satisfy the affordability criteria for your requested credit.} If a decision is made, we will inform you about the relevant information that has impacted the outcome.\\ \hline[dashed]
		E2.2: & 
		\newtcolorbox{graybox}{boxrule=0.5pt,colframe=black,colback=gray9,hbox,arc=0pt,left=0pt,right=0pt,top=0pt,bottom=0pt,nobeforeafter,fontupper={\small},box align=base,breakable,}
		{The system is calculating the probability of meeting our affordability criteria based on information about your financial standing that you provide in your application, that we already hold on file about you, or that we receive from our partner the CRA.\\
			\hlorange{You can experiment with how different financial information can impact our confidence that credit repayment is viable using the slider below. }\\
			\setlength{\fboxsep}{0pt}
			\begin{adjustbox}{width=.8\textwidth,center}
				\begin{tcolorbox}[title=Projected loan variations for responsible lending, center title,center,
					lifted shadow={1mm}{-2mm}{3mm}{0.1mm}%
					{black!50!white}, width=.9\columnwidth,enhanced,
					colback=lime!5!white,
					boxrule=0.1pt,
					colframe=lime!75!black,
					fonttitle=\bfseries,breakable,]
					
					\resizebox{\columnwidth}{!}{ \small
						\begin{tabular}{@{}p{3cm}@{} r|c|l@{}}
							Annual Income: & \textsterling 10,000.00 & \fbox{\aSlider{2.5cm}{0.5}} & \textsterling 100,000.00 \\ 
							Monthly instalment payment: & \textsterling 100.00 & \fbox{\aSlider{2.5cm}{0.3}} & \textsterling 50,000.00\\
							Previous Credit: & \textsterling 100.00 & \fbox{\aSlider{2.5cm}{0.1}} & \textsterling 50,000.00\\ 
							Permanent residency in the UK: & $<$ 1 year & \fbox{\aSlider{2.5cm}{0.7}} & $>$ 5+ years\\
							APR: & & \multicolumn{2}{c}{\radiobutton* Fixed \qquad \radiobutton Variable}\\
						\end{tabular}
					}
					
					\bigskip
					\hrule
					\bigskip
					For a requested credit amount \textsterling
					\begin{graybox}
						\texttt{2,500.00 \textemdash~4,500.00}
					\end{graybox} 
					a 
					\begin{graybox}
						\texttt{fixed}
					\end{graybox} 
					interest rate is projected to be 
					\begin{graybox}
						\texttt{3.8}
					\end{graybox}
					\%.
					\bigskip
					\hrule
					\bigskip		
					{\footnotesize \textit{Note: This tool is provided for illustration purposes only and does not constitute a decision or a promise of a decision of an actual application.}}
				\end{tcolorbox}
			\end{adjustbox}
			\vspace{-1em}
		}
		\\
		
		R3: & The \hlgreen{categories of personal data} used in making the decision	\\
		E3: & The data that will be processed includes \hlgreen{personal identifiers}, such as name, date of birth, and current and previous addresses; \hlgreen{information about your financial standing}, such as the number of credit and current accounts you hold and their repayment histories, salary data or insolvency-related events; \hlgreen{information about your credit rating}, such as scores about your credit, your affordability, or the possibility of insolvency or fraud.\\
		
		R4: & The \hlpurple{relevance} of the information for decision-making	\\
		E4: &  The personal identifiers are \hlpurple{used for matching}, so that we know that the processed data relates to you. Information about your finances and credit scores are \hlpurple{used to determine the likelihood that you will be able to repay your credit commitments}.\\
		
		R5: & {The \hlred{significance} of envisaged consequences} \\
		E5.1: & The screening process will determine \hlred{whether your application is successful}. \\
		E5.2: & {\hlred{If your application is successful, the Bank will sign a loan agreement with you} and
			credit you a loan of the amount specified in your application.
			\hlred{You will have to repay this loan} following the repayment conditions agreed in the loan agreement.
			Following our screening, \hlred{the Bank may consider that a loan of a different amount is financially viable to your specific circumstances}.
			\hlred{You will have the opportunity to accept or decline} the Bank's offer.
			The Bank will provide you with an explanation \hlred{in case your application is unsuccessful,
				explaining the reasons why}.
			You will have the chance to \hlred{challenge that decision if you consider that your application was not processed correctly}. }\\ 
	\end{longtblr}
}

Note that the typology can be used to create various types of explanations. For example, E2.2 uses a counterfactual explanation to illustrate the consequences of the underlying logic. Counterfactual explanations can demonstrate how different behaviour impacts the outcome of a process. However, their usefulness in placing the recipient in a position to challenge a financial decision that has already been made is limited since one is unlikely to be able to alter their financial assets (e.g.\ their income) at will.

Information given to the data subject after a decision has been made (\textit{ex post}) could expand the first set of explanations in the following way:

{\small 
	\begin{longtblr}[
		theme=fancy,
		caption={},
		label={none},
		]{colspec={|@{}l @{}X@{}|}, rowspec={|QQ| |QQ| |QQ| |QQ| |QQ|},
			row{1,3,5,7,Y}={bg=col2!90, font=\bfseries},}
		
		R1: & Decision taken solely by \hlblue{automated} means\\
		E1: & The Bank \hlblue{made this decision with an automated scoring system} that takes into account the information provided in the borrower application as well as a credit score produced by credit referencing agency the CRA\@. You can access the personal data about you that were processed for this decision by following \url{this}~\url{link}.\\
		
		R2: & \hl{Reasons} for the decision \\
		E2: & We regret to inform you that the loan application (\texttt{applications\_no/437}) was declined\@. \hl{This is because of negative credit history}.\\
		
        R3: & Key \hlgreen{data points} for the decision \\
        E3: & The borrower's credit score indicates the following negative information\@. \hlgreen{The late payment (\texttt{records/70551}), late payment (\texttt{records/70552}) and late payment (\texttt{records/70553}).} The assessment indicated that the commitment of the credit agreement \hlgreen{would have an adverse impact on the borrower's financial situation}. Hence the company decision (\texttt{applications\_no/437/decision}) is to refuse the borrower's application.\\
		
		R4: & \hlcyan{Process} to request a review of the decision \\
		E4: & If you are not satisfied with this decision, \hlcyan{you can get in touch} with us in the email address below. If you think some of the data in this decision are incorrect, you can request a review of the decision by one of our officers\@. \hlcyan{You will need to provide any new evidence you consider should be taken into account in assessing your application. To express your point of view or request a review, please send an email to \texttt{<email@thebank.bank>} quoting \texttt{applications\_no/437/decision}.}\\
		
		R5: & \hlorange{Response} to the review request \\
		E6: & \hlorange{We have reviewed decision \texttt{applications\_no/437/decision}} about you. In light of the new evidence you presented about late payment (\texttt{records/70551}) and late payment (\texttt{records/70552}), we are happy to inform you that we are in a position to \hlorange{offer you a credit amount of \texttt{application\_no/437/credit\_amount/adjusted}}. You have 30 days to respond if you would like to accept this offer. \\
	\end{longtblr}
}

Of note, Article 29 Working Party also includes within its own example, ``information to advise the data subject that the credit scoring methods used are regularly tested to ensure they remain fair, effective and unbiased''.~\citep{WP29:2018}. We however held the view that such a comforting statement did not mean much in practice if it was not backed up by additional information, which would allow data subjects to effectively verify the accuracy of such a claim. 

To assist the Bank's staff in monitoring and reviewing the decision-making process, we also produced the following explanations:
{\small
	\begin{longtblr}[
		theme=fancy,
		caption={},
		label={none},
		]{colspec={|@{}l @{}X@{}|}, rowspec={|QQ| |QQ| |QQ| |QQ| |QQ|},
			row{1,3,5,7,9}={bg=col2!90, font=\bfseries},}
		
		R1: & \hlgreen{Key decision points}\\
		E1: & The applicant's credit score is below the acceptance threshold of \texttt{750}. The applicant's credit score was impacted by \hlgreen{late payment (\texttt{records/70552}), missed payment (\texttt{records/70553}) and late payment (\texttt{records/70551})}. The source of this information was \hlgreen{credit history report \texttt{credit\_history/437}} provided by \texttt{credit\_agency\_3}. \\
		
		R2: & \hlred{Verification} of the results \\
		E2: & Identity matching was performed using \hlred{electoral register data received on \texttt{2021-01-03T02:00:00}}. The credit report and credit score from CRA were \hlred{calculated on \texttt{2020-10-03T04:00:00}}. Late payment (\texttt{records/70552}) of \texttt{2020-10-01T01:00:00} was \hlred{matched to late payment} in credit report \texttt{credit\_history/437}.\\
		
		R3: & \hlorange{Business rules} applied to the decision \\
		E3: & \hlorange{The threshold of acceptance} for \texttt{credit\_product/322} is \texttt{750}. The decision was balanced against the FCA rule about the Bank's \hlorange{commitment to responsible lending}. The decision took into account the protection of characteristics under the \hlorange{Equality Act 2010}.\\
		
		R4: & \hlcyan{Alternatives} to the decision \& reasons for \hlcyan{rejection} \\
		E4: & To overturn the decision, a score above \texttt{750} was necessary. The current score of \texttt{715} could justify a loan of up to £\texttt{10,000} with a fixed interest rate of \texttt{3/1} with an initial APR of \texttt{3.7\%}\@. \hlcyan{This option was rejected because the amount surpasses the acceptable difference of $\pm$\texttt{20}\%}.\\
		
		R5: & Review carried out by qualified \& authorised \hl{reviewer} \\
		E5: & The decision was \hl{reviewed at \texttt{2021-02-18T10:25:45.000Z} by credit officer (\texttt{staff/142})} who confirmed the credit and risk assessments.\\
	\end{longtblr}
}

These sample explanations are the result of a co-creation process, which implied signed-off by the industry partners. The explanations produced for internal monitoring and auditing are thus richer (in terms of details) than the explanations addressed to data subjects. It is likely that the range of explanations addressed to data subjects would need to be expanded to include additional information, such as information about the respective weights of the factors taken into account to produce the decision, i.e., when a machine learning model is deployed the respective weights of the relevant model features~\cite{Ausloos:2025}, which we could not capture with the provenance trail developed for the purpose of this paper. In any case, to determine whether the information provided meets an explanation requirement, the information should be assessed in the light of the explainability goal pursued. If contestability is a relevant goal, then it is not sufficient to show that the addressee of the decision is in a position to make sense of the information provided. The decision-maker should also make sure that the information provided enables the addressee of the decision to assess the lawfulness of the decision-making process and its output.  

\subsection{School Allocation}\label{subsection: school allocation}

\begin{sidewaystable}
	\resizebox{\textwidth}{!}{
		\begin{talltblr}[
			caption = {},
			label = {none},
			]{
				colspec = {@{}p{5cm}llllllllllllp{4cm}llllllll@{}}, width = \linewidth,
				row{odd} = {white}, row{even} = {red!45!white!15},
				row{1-3} = {green!50!black!15,font=\bfseries},
				hline{4,Z} = {1pt},
			}
			
			& \SetCell[r=1]{f}Source & \SetCell[c=2]{c}{\rotatebox{60}{Timing}} && \SetCell[c=2]{c}{\rotatebox{60}{Autonomy}} && \SetCell[c=3,r=1]{c,f}{Trigger} &&& \SetCell[c=5,r=1]{c,f}{Content} &&&&& \SetCell[c=2]{c}{\rotatebox{60}{Scope}} && \SetCell[c=2,r=1]{c,f}{Explainability goal} && \SetCell[c=2,r=1]{c,f}{Intended recipient} && \SetCell[c=2]{c}{\rotatebox{60}{Criticality}} \\ \cmidrule[lr]{3-4} \cmidrule[lr]{5-6} \cmidrule[lr]{7-9} \cmidrule[lr]{10-14} \cmidrule[lr]{15-16} \cmidrule[lr]{17-18} \cmidrule[lr]{19-20} \cmidrule[lr]{21-22} 
			&  & \rotatebox[origin=r]{90}{Ex ante} & \rotatebox[origin=r]{90}{Ex post} & \rotatebox[origin=r]{90}{Proactive} & \rotatebox[origin=r]{90}{Reactive} & \rotatebox[origin=r]{90}{Decision}  & \rotatebox[origin=r]{90}{Process} & \rotatebox[origin=r]{90}{Action} & \SetCell[c=2]{c}{\rotatebox[origin=r]{90}{Sensitivity}} && \SetCell[c=2]{c}{\rotatebox[origin=r]{90}{Confidentiality}} && Minimum Content & \rotatebox[origin=r]{90}{Universal} & \rotatebox[origin=r]{90}{Local} & Understandability & Intervenability & Outward-facing & Inward-facing & \rotatebox[origin=r]{90}{Mandatory} & \rotatebox[origin=r]{90}{Recommended} \\ \cmidrule[lr]{10-11}\cmidrule[lr]{12-13}
			&  &  &  &  &  &   &  &  & \rotatebox[origin=r]{90}{\parbox{2.1cm}{Non-Directly Identifying}} & \rotatebox[origin=r]{90}{\parbox{2.1cm}{Directly Identifying}} & \rotatebox[origin=r]{90}{\parbox{2.1cm}{Disclosed}} & \rotatebox[origin=r]{90}{\parbox{2.1cm}{Restricted}} & &  &  &  &  &  &  &  &  \\
			
			Reasons for the decision & Primary & & $\bullet$ & & $\bullet$ & $\bullet$  & &  & & $\bullet$ &$\bullet$& & Scoring on school admission criteria per school & & $\bullet$  & {Justification \\Fairness} & Contestability & Decision Addressee &  & $\bullet$ &  \\
			
			Information related to the right to appeal & Primary & & $\bullet$ & & $\bullet$ & $\bullet$  & &  & $\bullet$ & &$\bullet$& & {Right \\ Timeline \\ Procedure}  & $\bullet$ &  & {Accountability \\ Consequence} & {Contestability} & Decision Addressee &  & $\bullet$ & \\
			
			Consequences of the decision & Secondary & & $\bullet$ & & $\bullet$ & $\bullet$  & &  & $\bullet$ & &$\bullet$& & {Consequences attached to the decision related to waiting lists}  & $\bullet$ &  & {Consequences \\ Fairness} & Contestability & Decision Addressee &  & & $\bullet$  \\
			
		\end{talltblr}
	}
	\caption{Classification of explanation requirements related to the School Admissions Code}\label{tab:SA}
\end{sidewaystable}

In this scenario, Bob and Mary live in Southampton, UK\@. They have a son Chris, who will turn 4 years old in May. He should thus start school in September.
Bob and Mary apply for a school place at primary school level.
As part of the application process, Bob and Mary declare their preferences for three schools: \emph{Bassett Green Primary}, \emph{Oakwood Primary} and \emph{St Denys Primary School}.
In June, they receive a letter from the Southampton city council telling them that Chris has been admitted at \emph{Oakwood Primary}.  

In England, school admissions are covered by the School Admissions Code 2021~\cite{DepartmentEducation:2021}.
The Admissions Code is a statutory code of practice issued under the School Standards and Framework Act 1998 (SSFA).\negmedspace\footnote{School Standards and Framework Act 1998 c. 31, ss. 84 and 85.}
All admission authorities and admission appeal panels have a statutory duty to comply with it.\negmedspace\footnote{School Standards and Framework Act 1998 c. 31 s. 84.}  

Each school has discretion to choose oversubscription criteria that prioritise applicants if the applications are more than the number of available places. When assessing the application for a child, the local authority must apply the oversubscription criteria of each of the preferred schools. If the child satisfies the oversubscription criteria of more than one schools, a place at the highest ranked of these school is offered. If the child does not satisfy the oversubscription criteria of any of the preferred schools, the Council must offer the child a place at the nearest school with empty places. 

In our example, \emph{Basset Green Primary} prioritises applicants according to five criteria:  
\begin{enumerate}
    \item Children in public care
    \item Children subject to a child protection plan
    \item Children whose sibling is on the roll of the school
    \item Children who attend one of the feeder nurseries
    \item Children living closest to the school 
\end{enumerate}

\emph{Oakwood Primary} prioritises applicants according to 3 criteria:  

\begin{enumerate}
    \item Children in public care
    \item Children whose sibling is on the roll of the school
    \item Children living closest to the school 
\end{enumerate}

In addition, since admission processes require the processing of personal data, the Data Protection Act 2018 (DPA 2018) and the UK GDPR apply, and in particular UK GDPR Art. 12 and 13. Article 22 is out of scope as long as the process is not merely automated.  

For the purposes of this paper, we focus upon two requirements stemming from the School Admissions Code. Para 2.32 of the Code imposes upon admission authorities the following requirement:  

\textit{``When an admission authority informs a parent of a decision to refuse their child a place at a school for which they have applied, it must include the reason why admission was refused; information about the right to appeal; the deadline for lodging an appeal and the contact details for making an appeal. Parents must be informed that, if they wish to appeal, they must set out their grounds for appeal in writing. Admission authorities must not limit the grounds on which appeals can be made.''}

Such a requirement can be broken down into two requirements as illustrated in \cref{tab:SA}. 

Besides, for the sake of completeness, the Southampton school admission team decided to add a requirement to explain the consequences of accepting the decision.

To assist the City Council in producing explanations once school-allocation decisions have been made, we produced the following examples of explanations. 
The first example aims to inform parents/guardians of the reasons for the decision, highlighting the decision criteria per school and disclosing the breakdown of children meeting each criterion.  
The second example aims to inform parents/guardians of the next steps they can take once they have received the decision and want to appeal the decision.
The third examples aims to inform parents/guardians about the consequences of accepting the decision.  
The minimum content is highlighted for each requirement. 

{\small
	\begin{longtblr}[theme=fancy,caption={},label={none}]{colspec={|@{}l @{}X@{}|}, rowspec={|QQ| |QQ| |QQ|},row{1,3,5}={bg=col2!90, font=\bfseries},measure=vbox}
		
		R1: & \hlgreen{Reasons for the decision}\\
		E1: & Too many applicants applied for a place in \textbf{Bassett Green Primary} and as a result we are not able to offer Chris a place in this school.
        \textbf{Bassett Green Primary} prioritises applicants according to five criteria: Crit\#1, Crit\#2, Crit\#3, Crit\#4, and Crit\#5. Chris satisfied criterion Crit\#5, \hlgreen{where he was ranked at \textbf{Rank\#33}} within Crit\#5 based on the distance of your house to the school. Unfortunately, given the number of applicants satisfying prior criteria, admission of children under criterion Crit\#5 was \hlgreen{cut off at \textbf{Rank\#18}}. Any further admission to this year group will prejudice the provision of efficient education and the efficient use of resources. 
        A place has been allocated to Chris at \textbf{Oakwood Primary}, your 2nd preference school, starting September 2024. 
        
        A detailed breakdown of admitted children by admission criterion can be found below.
        
        \textbf{Basset Green Primary}:  
        \begin{itemize}
            \item Published Admission Number: \textbf{175}
            \item Application received: \textbf{616}
        \end{itemize}
        \begin{description}
            \item[Crit\#1] Children in public care \textbf{15}
            \item[Crit\#2] Children subject to a child protection plan \textbf{17}
            \item[Crit\#3] Children whose sibling is on the roll of the school \textbf{30}
            \item[Crit\#4] Children who attend one of the feeder nurseries \textbf{105}
            \item[Crit\#5] Children living closest to the school \textbf{18}
        \end{description}
        The last child to be admitted at \textbf{Basset Green Primary} under Crit\#5 was ranked 175 out of 616 on their list of applicants and lived 1.9 miles from the school.

        \textbf{Oakwood Primary}:
        \begin{itemize}
            \item Published Admission Number: \textbf{180}
            \item Application received: \textbf{432}
        \end{itemize}
        \begin{description}
            \item[Crit\#1] Children in public care \textbf{8}
            \item[Crit\#2] Children whose sibling is on the roll of the school \textbf{85}
            \item[Crit\#3] Children living closest to the school \textbf{87}
        \end{description}
        The last child to be admitted at \textbf{Oakwood Primary} under Crit\#3 was ranked 180 out of 432 on their list of applicants and lived 2.8  miles from the school. \\
        
		R2: & \hlred{Information} about the right to appeal \\
		E2: & If you wish to appeal the Council’s decision, you must lodge your appeal by \hlred{30 June 2024} at \hlred{appeal@southampton.gov.uk}. You can find more information on how to commence the process at http://www.southampton.gov.uk/schoolappeals. You should specify your grounds \hlred{in writing}.\\
		
		R3: & \hlorange{Consequences} of the decision \\
		E3: & If you \hlorange{accept} the place at Oakwood Primary your child’s name \hlorange{will still go on the waiting list for your higher ranked school}, i.e., Bassett Green Primary.\\
	\end{longtblr}
}

\section{Evaluation}\label{section:evaluation}

Once explanation requirements were generated in collaboration with domain experts from our industry partners, explanation outputs were assessed by representatives of producers and consumers of explanations gathered by our industry partners. We then revisited our methodology and typology in the light of the EU AI Act adopted later on.

\subsection{Prior to the AI Act}
The pilot studies, conducted prior to the adoption of the EU AI Act, confirmed the feasibility and usefulness of the methodology and typology that was produced for the purpose of computing explanations. By feasibility, we refer to the practicality of translating high-level explanation requirements found in regulatory, compliance or business mandates into specific, computable components. This involved testing whether high-level explanation requirements could be deconstructed into smaller, actionable parts and whether these could be processed by a computational system to provide outputs to users. The study allowed us to explore related operational challenges, such as challenges related to the gathering and translation of explanation requirements and the identification of relevant human and computational resources.

Provenance data models make it possible to traverse and query audit trails of the processing, 
records that describe the people, institutions, entities, and activities involved in producing, influencing, or delivering the outcome of a processing operation~\cite{Moreau:2013}.
We thus designed a provenance data model capable of inspecting the data flows generated by an automated loan application process. In a nutshell, the Minimum Content category was translated into a provenance pattern that retrieves the necessary data from the audit trail, while the Intended Recipient and Explainability Goal were used by a Natural Language Generation engine to construct the explanation after a linguistic syntax tree. 

In addition to feasibility, the study also examined the usefulness of these explanations, meaning how well they serve their intended goals. A useful explanation must not only be technically feasible but also meaningful for its audience. This includes testing whether the generated explanations allowed users to get benefits from the output in the light of the state of the art, i.e., existing practices. 

The feasibility was demonstrated through the production of a light ontology and an explanation assistant, which was created to address the needs of two use cases: credit scoring and school admissions. The light ontology can be found in Appendix A. Through a machine-readable ontology, we adopted a systematic terminology for concepts, reflected on their inter-relations, and derived hierarchical representation.
It also allowed us to systematically encode sets of requirements according to the ontology. These sets of requirements are subsequently used in the explainability by design methodology to generate explanations.

The Explanation Assistant is a software service configured to support a set of explanation requirements generated by a Legal Engineer following the proposed taxonomy. From the application data associated with a decision, it produces narratives in a natural language to explain that decision. Example of computable explanation narratives were presented to two industry partners in the context of interactive focus groups: one CRA (Experian UK) and one school admission team (from Southampton City Council).\negmedspace\footnote{For more information, see \url{https://plead-project.org/research/}.} 

As regards usefulness, producers (data teams) and consumers (teams in charge of responding to end-user requests for information) of explanations from our industry partners were then asked to assess the explanations, taking into account the solutions that they had already deployed prior to the beginning of the project. They were thus gathered in two focus groups to answer a set of questions in a moderated setting. Each group was chosen due to its prior interactions with explanations produced by each industry partner, and the questions were designed to shed light on the quality of explanation outputs, taking into account the two explainability goals mentioned above: intervenability and understandability. All participants to the focus groups confirmed that the explanations produced on the basis of the typology went further than their existing practices and that if implemented would improve these practices by enhancing the level of understandability, either because they included more details about the decision-making process or because they included more tailored information for the recipients of the explanations.
During interactions with producers and consumers of explanations, participants made clear their intention either to become more sophisticated providers of explanations themselves or to delegate explanation-related tasks to third parties with appropriate expertise. 

The benefit of the typology was considered to be twofold:
First, it forces the legal engineer to systematically identify all building blocks needed to compute requirements for explanation automation. Second, the typology helped the legal engineer to systematically organise explanation requirements and identify duplicates. An organisation may find that multiple explanation requirements map to the same dimensions and sub-categories. These requirements can therefore be addressed by similar explanation(s), allowing the organisation to scale their compliance strategies. The legal engineer was then ready to be embedded within the engineering team to help it develop a software architecture relying upon reusable components and aiming to incorporate the resulting explanation capability into an application. While the fine-grained explanation requirements generated on the basis of the typology have been used as input for the work of the data engineer, an iterative process supported by an on-going dialog between the legal engineer and the data engineer had to be followed to perform a variety of engineering tasks, such as modelling the decision audit trail, creating queries to extract data from those audit trails for each explanation, building explanation plans to transform the queried data in narratives consumable by end-users, and ultimately deploying the Explanation Assistant (see~\cite{Huynh:2021} for more information).

Considering these benefits in the light of prior work and in particular prior review of methods and techniques that are being developed to improve explainability in machine learning prediction tasks~\cite{Zhou:2021}, it is possible to argue that the typology can be used as a means to assess the quality of the explanatory approach when the ML model is integrated within a wider decision-making pipeline and to choose the most appropriate explanation for a given explanation requirement, be it a mere business requirement or a business requirement informed by applicable law.
The typology forces the elicitation of explanation requirements, which are then used as benchmarks to tailor the content of the explanation, addressing all dimensions of the typology, in an attempt to reduce subjectivity. This process should be useful to make hidden cognitive biases and social expectations vis-à-vis the explanation process~\cite{Miller2017} emerge more clearly.

Reflecting upon the research and innovation process implemented during the pilot studies, the main characteristics of the legal engineer role emerged more clearly and are listed below:
\begin{itemize}
  \item Role overview: This interdisciplinary role is responsible for translating legal principles and compliance standards into secondary requirements that are actionable by the engineering team. 
  \item Key responsibility: First, this role must gather high-level requirements from the compliance and legal team and deliver a preliminary set of secondary actionable requirements to the engineering team. Second, while embedded within the engineering team, the role holder must iteratively refine the preliminary set of requirements through interactions with the engineering team on a regular basis to support actively the production of a working prototype. 
  \item Collaboration and communication: This role must be in regular communication with two teams, i.e., the legal and compliance team as well as the engineering team.
  \item Skill and qualification: this role must have a legal background and a solid understanding of technology and engineering principles. While not necessarily requiring a formal degree in a technical field, relevant certifications or hands-on experience in data science, computer science, or information technology are advantageous. 
  \item Reporting structure: Working at the interface of a legal and compliance team and an engineering team the reporting structure can vary depending on the organisation's size, structure, and priorities. However, one option in particular would ensure that the legal engineer's expertise is appropriately leveraged. It is recommended that the legal engineer report to a Data Governance Officer, whose responsibilities lie in ensuring that data-related activities adhere to legal and regulatory standards and whose role is not merely advisory. 
  \item Dependencies: This role is characterised by dependencies on effective communication and cooperation with two teams. First, the legal and compliance team, with whom the legal engineer must work closely to determine the range of applicable legal frameworks and associated internal compliance standards, the interpretative position of the organisation as regards key legal concepts and ethical commitments; and second, with the engineering team, with whom the legal engineer must work closely to understand technical requirements, constraints, and opportunities for embedding compliance measures in data systems or pipelines.
\end{itemize}

\subsection{The AI Act}\label{subsection: AI Act}

\begin{sidewaystable}
	\resizebox{\textwidth}{!}{
		\begin{talltblr}[
			caption = {},
			label = {none},
			]{
				colspec = {@{}p{5cm}llllllllllllp{4cm}llllllll@{}}, width = \linewidth,
				row{odd} = {white}, row{even} = {red!45!white!15},
				row{1-3} = {green!50!black!15,font=\bfseries},
				hline{4,Z} = {1pt},
			}
			
			& \SetCell[r=1]{f}Source & \SetCell[c=2]{c}{\rotatebox{60}{Timing}} && \SetCell[c=2]{c}{\rotatebox{60}{Autonomy}} && \SetCell[c=3,r=1]{c,f}{Trigger} &&& \SetCell[c=5,r=1]{c,f}{Content} &&&&& \SetCell[c=2]{c}{\rotatebox{60}{Scope}} && \SetCell[c=2,r=1]{c,f}{Explainability goal} && \SetCell[c=2,r=1]{c,f}{Intended recipient} && \SetCell[c=2]{c}{\rotatebox{60}{Criticality}} \\ \cmidrule[lr]{3-4} \cmidrule[lr]{5-6} \cmidrule[lr]{7-9} \cmidrule[lr]{10-14} \cmidrule[lr]{15-16} \cmidrule[lr]{17-18} \cmidrule[lr]{19-20} \cmidrule[lr]{21-22} 
			&  & \rotatebox[origin=r]{90}{Ex ante} & \rotatebox[origin=r]{90}{Ex post} & \rotatebox[origin=r]{90}{Proactive} & \rotatebox[origin=r]{90}{Reactive} & \rotatebox[origin=r]{90}{Decision}  & \rotatebox[origin=r]{90}{Process} & \rotatebox[origin=r]{90}{Action} & \SetCell[c=2]{c}{\rotatebox[origin=r]{90}{Sensitivity}} && \SetCell[c=2]{c}{\rotatebox[origin=r]{90}{Confidentiality}} && Minimum Content & \rotatebox[origin=r]{90}{Universal} & \rotatebox[origin=r]{90}{Local} & Understandability & Intervenability & Outward-facing & Inward-facing & \rotatebox[origin=r]{90}{Mandatory} & \rotatebox[origin=r]{90}{Recommended} \\ \cmidrule[lr]{10-11}\cmidrule[lr]{12-13}
			&  &  &  &  &  &   &  &  & \rotatebox[origin=r]{90}{\parbox{2.1cm}{Non-Directly Identifying}} & \rotatebox[origin=r]{90}{\parbox{2.1cm}{Directly Identifying}} & \rotatebox[origin=r]{90}{\parbox{2.1cm}{Disclosed}} & \rotatebox[origin=r]{90}{\parbox{2.1cm}{Restricted}} & &  &  &  &  &  &  &  &  \\
			
			Origin of the data & Primary:implicit & &$\bullet$& $\bullet$ & $\bullet$ &   & $\bullet$ &  & $\bullet$ &  & &$\bullet$& Source, e.g., data provider, data subject & $\bullet$ &  & {Accountability \\ Accuracy \\ Performance} & {Scrutability} & & {Administrator \\ Data Engineer \\ Auditor} & $\bullet$& \\
			
		  Data collection process & Primary:implicit & &$\bullet$& $\bullet$ & 
            $\bullet$ &   & $\bullet$ &  & $\bullet$ &  & &$\bullet$& E.g., web scraping, API access,\ldots & $\bullet$ &  & {Accountability \\ Overall Transparency} & {Scrutability}  & & {Administrator \\ Auditor} &$\bullet$ &   \\
			
			Data collection purpose & Primary:implicit & &$\bullet$& $\bullet$ & $\bullet$ &   & $\bullet$ &  & $\bullet$ &  & &$\bullet$& Phase of model lifecycle (e.g., training, deploymnent) & $\bullet$ &  & {Accountability \\ Data Minimisation} & Scrutability  & & {Administrator \\ Auditor} & $\bullet$&  \\
			
			Data preparation actions & Primary: implicit & &$\bullet$& $\bullet$ & 
            $\bullet$ &   & $\bullet$ &  & $\bullet$ &  & &$\bullet$& Data curation actions & $\bullet$ &  & {Accountability \\ Accuracy \\ Performance} & Scrutability & & {Administrator \\ Data engineer \\ Auditor} &$\bullet$ &  \\
			
			Bias mitigation actions & Primary:implicit & &$\bullet$& $\bullet$ & 
            $\bullet$ &   & $\bullet$ &  & $\bullet$ &  &$\bullet$& & Mitigation echniques & $\bullet$ &  & {Accountability \\ Performance \\ Fairness} & Scrutability & & {Administrator \\ Data engineer \\ Auditor} &$\bullet$ &  \\
			
			Oversight actions & Primary:implicit & &$\bullet$& $\bullet$ & 
            $\bullet$ &   & $\bullet$ &  &  & $\bullet$ &$\bullet$& & Implementation of explainability techniques et processes & $\bullet$ &  & {Accountability \\ Overall Transparency} & Scrutability & & {Administrator \\ Auditor} & $\bullet$ &  \\
			
			Role of the AI system & Primary & &$\bullet$& &$\bullet$& $\bullet$ & &  & $\bullet$ & &$\bullet$& &  & $\bullet$ & {E.g., production of information summaries \\ sorting \\ recommending \\ access grants} & {Accountability \\ Overall Transparency \\ Fairness} & Contestability & Decision Addressee & & $\bullet$ &  \\

            Meaningful elements of the decision & Primary & &$\bullet$& &$\bullet$& $\bullet$ & &  & &$\bullet$ & &$\bullet$ & {Elements meaningful to contest the decision \\ (e.g., input, source of input, features, consequence of decision)} & &$\bullet$& {Justification \\ Fairness} & {Contestability \\Human intervention} & Decision Addressee &  & $\bullet$ &  \\
		\end{talltblr}
	}
	\caption{Examples of explanation requirements derived from the AI Act}\label{tab:AIAct-1}
\end{sidewaystable}

The EU AI Act is a horizontal regulation recently finalised by the EU, which prohibits a certain number of AI practices,\negmedspace\footnote{AI Act,  Article 5.} governs the placing on the market and the putting into service of high-risk AI systems\footnote{AI Act, Chapter III.} and general-purpose AI models and Systems,\negmedspace\footnote{AI Act, Articles 53--56.} and introduces transparency obligations for certain AI systems.\negmedspace\footnote{AI Act, Article 50.} 
As regards high-risk AI systems, in which category one finds ``AI systems intended to be used to evaluate the creditworthiness of natural persons or
establish their credit score, with the exception of AI systems used for the purpose of
detecting financial fraud''\footnote{AI Act, Annex III 5(b)} and ``AI systems intended to be used to determine access or admission or to assign natural
persons to educational and vocational training institutions at all levels,''\footnote{AI Act, Annex III 3(a)} Chapter III is of particular relevance. Within this chapter, Article 10 imposes a whole set of data governance and management practices, which include for example tracking: 
\begin{itemize}
    \item ``data collection processes and the origin of data, and in the case of personal data, the original purpose of the data collection;''\footnote{AI Act, Article 10(2)(b).}
    \item ``relevant data-preparation processing operations, such as annotation, labelling, cleaning, updating, enrichment and aggregation;''\footnote{AI Act, Article 10(2)(c).}
    \item ``appropriate measures to detect, prevent and mitigate possible biases identified.''\footnote{AI Act, Article 10(2)(g).}
\end{itemize}
Article 10 also introduces a data relevancy requirement.\negmedspace\footnote{AI Act, Article 10(3).}
Article 12 introduces a record-keeping obligation and Article 15 a requirement for effective human oversight. 
These provisions thus contain substantive requirements which could be evidenced through the generation of explanations, to the primary benefits of internal roles, but also third-party auditors and eventually regulators. 
In addition, Article 86 introduces a right to explanation to the benefits of any affected person subject to a decision which is taken by the deployer on the basis of the output from a high-risk AI system when two sets of conditions are met: legal or similar effects plus in-scope adverse effects must be produced. When these conditions are met, the affected person has the right to obtain an explanations of the role of the AI system in the decision-making procedure and the main elements of the decision taken.

Following our methodology and leveraging our typology we could for example produce a set of explanations illustrated in \cref{tab:AIAct-1}. 
The last requirement in the table will have to be broken down in more granular sub-requirements.

\section{Limitations}\label{section:limitation}

All in all, even if the pilot studies achieved their goal, a comprehensive evaluation process remains needed. The pilot studies included stakeholders from the finance and education sectors only. Although the typology has been designed to be domain-agnostic, evaluation in other sectors would be useful to test the comprehensiveness of the typology and refine or adapt its dimensions and sub-categories for other verticals. More work is also needed to carefully derive explanation requirements from all relevant obligations found in the EU AI Act. 

Importantly, despite the highly integrated process that was followed to produce explanation requirements through leveraging the typology, the process remains a manual one, which cannot be fully automated, particularly when the process starts with a translation of primary requirements stemming from applicable laws. Hence, the importance of the role of the Legal Engineer, who will have to interpret high-level requirements and transform them into actionable requirements, while acknowledging possible reductions. Among the different dimensions and sub-dimensions, the Minimum Content one, in particular, requires careful analysis of the high-level requirements, which will then have to be assessed in the light of technical capabilities. Crucially, it is worth noting that computable fine-grained explanation requirements, which can be opinionated, do not necessarily exhaust higher level requirements.
By way of example, the provenance trail that may be generated to achieve accuracy is likely to be focused upon integrity, yet data subjects have a right under GDPR Article 15 to access data to ensure that their personal data is correct as well as to verify the lawfulness of the processing.\negmedspace\footnote{See e.g., CJEU C-487/21 FF v Österreichische Datenschutzbehörde and CRIF GmbH, para. 34.}
In any case, although this research was constrained by the co-creation process itself, it demonstrated that refusals to explain underlying data processing based on the complexity of the processing are not justified and that a uniform approach can be adopted for both purely and partially automated decision-making pipelines.

More generally, an organisation wanting to operationalise the production of explanations based on the typology introduced in this paper should carefully map applicable explanation requirements to the nine dimensions of the typology. As a result, the quality of the classification will largely depend on the quality of the mapping. 

What is more, the quality of the explanations produced on the basis of the typology will also depend upon the quality of the description of the decision-making pipeline and its output. By way of example, the organisation implementing the typology will want to ensure it is able to distinguish between public and restricted explanations.  In other words, determining when an explanation can or should be disclosed to third parties is a decision for the implementing organisation to be made on the basis of an internal classification.
As mentioned in \cref{section:discussion}, the GDPR allows data controllers to carefully balance the amount of information they disclose to individuals with the protection of their commercial interests as long as they do not undermine data subject rights.\negmedspace\footnote{See GDPR Article 15(4).}
In these cases, it is up to the implementing organisation to decide which information should be included in an explanation and how best to use the Confidentiality sub-category of the Content dimension to control disclosure.

Another takeaway relates to the necessity to work with rich input to be able to provide actionable output~\cite{Solano:2019}. 

Finally, a significant challenge in the context of explanation computation is trustworthiness. It is essential to define metrics that can be used to assess the integrity of these explanations and ensure they have not been tampered with.

\section{Conclusion}\label{section:conclusion}

This paper presented a typology built to categorise explanation requirements that can be leveraged to compute explanations. The typology comprises nine dimensions, with each dimension forming a set of building blocks for the generation of explanations. This typology is the product of both conceptual and empirical research and has been co-created with industry partners, who are both consumers and producers of explanations. The feasibility of its construction for the purpose of computing explanations has been tested in the context of two small-scale pilot studies. The usefulness of the outputs produced on the basis of the typology for producers and consumers operating within the context of these use cases has also been tested. While \cref{section:typology} unpacks the nine dimensions of the typology, \cref{section:discussion} describes the co-creation process that was followed and includes examples of explanations produced in the context of two scenarios\@. \cref{section:evaluation} discusses the evaluation of the pilot studies and the importance of a new role, that of the legal engineer, of which main characteristics are broken down and further specified. Driven by a responsible innovation approach, we found that adding an interdisciplinary role, that of the legal engineer, was an effective way to bridge the work of the legal, compliance, and engineering teams and, crucially, acknowledge the limits of automation. The legal engineer was responsible for translating high-level requirements received from legal and compliance teams into fine-grained explanation requirements to be consumed by the engineering team. The legal engineer was subsequently embedded within the engineering team and regularly consulted during the technical design phase to help the data engineer specify the requirements for the software architecture components.

The research led to the development of a light vocabulary based on the typology, included in Appendix A and an explanation assistant, a software service configured to support a set of explanation requirements. The light vocabulary provides a machine-readable format for classifying fine-grained explanation requirements.
It can be combined with existing compliance or auditing ontologies, such as, for example, the consent ontology developed by \citet{Pandit:2019}.

Our next steps will be to perform a comprehensive evaluation process and add new use cases to the two use cases mentioned above as well as report more extensively on the technical design phase.

\section{Acknowledgments}

The work presented in this article has been supported by the UK Engineering and Physical Sciences Research Council (EPSRC) via the Grants [EP/S027238/1] and [EP/S027254/1] for the PLEAD project. The authors have no competing interests to declare that are relevant to the content of this article.

\bibliographystyle{ACM-Reference-Format}
\bibliography{manuscript}

\newpage

\section*{Appendix A}
\begin{figure}[!htb]
	\centering
	\begin{minipage}[b]{.5\linewidth}
		\includegraphics[width=.8\linewidth]{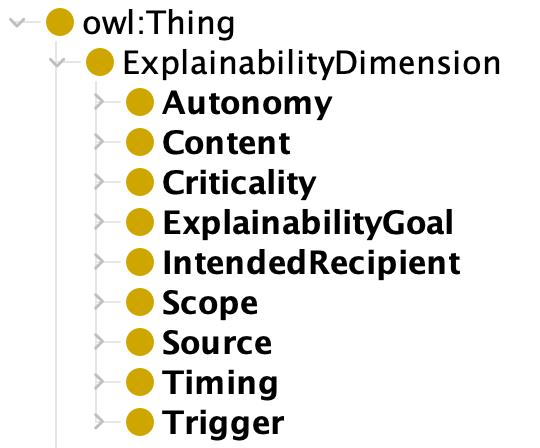}
		\captionof{figure}{Top level classes of PLEAD Explainability-by-Design typology}\label{fig:classes}
	\end{minipage}\hfill
	\begin{minipage}[b]{.5\linewidth}
		\includegraphics[width=.8\linewidth]{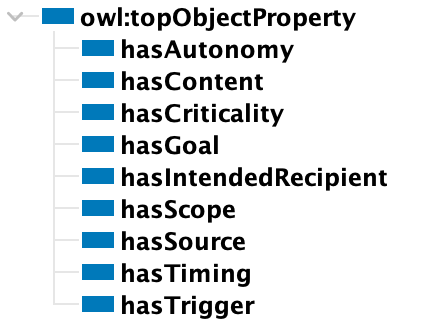}
		\captionof{figure}{Object properties of PLEAD Explainability-by-Design typology}\label{fig:objects}
	\end{minipage}
\end{figure}

The Explainability-by-Design typology is encoded as a lightweight vocabulary encoded as an \textsc{owl}{\small 2} ontology.  
Its namespace is \url{https://openprovenance.org/ns/plead}.
It is designed to be as simple as possible, consisting of 10 top-level classes and 9 object properties. 
The 10 classes model the concept of Explanation and its associated 9 dimensions (\cref{fig:classes}). 
The 9 object properties have Explanation as their domain and the corresponding dimension class as their range (\cref{fig:objects}).

An instance of explanation (such as Existence of automated decision-making in \cref{tab:GDPR22-1}) can be expressed as set of RDF statements, as in Listing~\ref{lst:rdf}.

\begin{lstlisting}[basicstyle=\LSTfont, columns=fullflexible, xleftmargin=5mm, framexleftmargin=5mm, numbers=left, stepnumber=1, breaklines=true, breakatwhitespace=false, numberstyle=\footnotesize, numbersep=5pt, tabsize=2, frame=lines, captionpos=b, caption={Explanation instance in RDF},label={lst:rdf},keepspaces,]
	###  https://openprovenance.org/ns/plead#xplain1
	plead:xplain1 rdf:type owl:NamedIndividual ,		
		plead:ExistenceAutomatedDecisionMaking ;
			plead:hasAutonomy plead:proactive ;
			plead:hasContent plead:generalised ,
												plead:disclosed ,
												plead:minimum1 ;
			plead:hasGoal plead:fairness1 ,
										plead:accountability1 ,
										plead:overall_transparency1 ;
			plead:hasIntendedRecipient plead:consumer1 ;
			plead:hasTiming plead:ex_ante ;
			plead:hasCriticality plead:mandatory ;
			plead:hasScope plead:universal ;
			plead:hasSource plead:implicit_art_22 ;
			plead:hasTrigger plead:loan_application ;
		plead:example "To be able to process applications quickly and 
										accurately, the Bank uses an automated system." .
\end{lstlisting}

\section*{Appendix B}\label{appendix:B}
\Cref{tab:examples} is listing examples out of the pilot studies, where the study participants validated proposed explanations for the two use cases. The excerpts are correlated in the table to their assigned `base', i.e.\ identified themes that were then used to form explanation requirements. 

\begin{table}[!htb]
	\begin{tabular}{@{}lp{.3\textwidth}p{.6\textwidth}@{}}
		\toprule
		ID   & Requirement                                                            & Example excerpt(s)                                                                                                                                                                                                                                                                                                                                                                                                                                                                                                                                                                                                                                                                                                                                                                                                                                                                                                              \\ \midrule
		RQ1  & Explanations must comply with a legal obligation                       & ``Yeah, well {[}\ldots{]} we provide the query process {[}because{]} there is effectively law that says the bank has to tell the customer which CRA was used.''                                                                                                                                                                                                                                                                                                                                                                                                                                                                                                                                                                                                                                                                                                                                                                      \\
		RQ2  & Explanations must comply with company policy                           & \begin{tabular}[c]{@{}p{.6\textwidth}@{}}``Now that CRA data goes into the bank and they make a decision based on their policy, so surely you need that angle with some nicely worded {[}explanation{]}''\\ ``For me, you're quite right the policy piece is important''\end{tabular}                                                                                                                                                                                                                                                                                                                                                                                                                                                                                                                                                                                                                                                         \\
		RQ3  & Explanations must address different contexts                           & ``I think that would be really, really handy for our contact centre as well.''                                                                                                                                                                                                                                                                                                                                                                                                                                                                                                                                                                                                                                                                                                                                                                                                                                                  \\
		RQ4  & Explanations must be scalable and re-usable                            & ``I think it would minimize the amount of traffic coming through {[}the call centre{]} as well''                                                                                                                                                                                                                                                                                                                                                                                                                                                                                                                                                                                                                                                                                                                                                                                                                                \\
		RQ5  & Explanations must be produced regardless of data subject participation & ``we kind of don’t have to dig for that information unless obviously you know someone creates {[}a request{]}. We wouldn't actually look into {[}the explanation{]} ourselves. So having {[}the explanation{]} there would be quite useful.''                                                                                                                                                                                                                                                                                                                                                                                                                                                                                                                                                                                                                                                                                   \\
		RQ6  & Explanations must convey the necessary information fast                & \begin{tabular}[c]{@{}p{.6\textwidth}@{}}``\ldots as members of the team {[}we{]} wouldn't actually be able to have that to hand necessarily {[}w/out the explanation{]}''\\ ``I think it would minimize the amount of traffic coming through {[}the call centre{]} as well.''\end{tabular}                                                                                                                                                                                                                                                                                                                                                                                                                                                                                                                                                                                                                                                         \\
		RQ7  & Explanations must empower the recipient                                & ``Definitely adds value. Like we said, it's kind of putting the ownership back into the hands of the consumer so that they can actually see physically.''                                                                                                                                                                                                                                                                                                                                                                                                                                                                                                                                                                                                                                                                                                                                                                       \\
		RQ8  & Explanations must educate the recipient                                & \begin{tabular}[c]{@{}p{.6\textwidth}@{}}``It would actually be accessible to them {[}the data subjects{]} quite quickly, and so it would probably alleviate some of the calls that we would be getting, and it would explain in more detail exactly how that decision was made.''\\ ``it kind of comes back to your tool, which I quite like because then that kind of gives that extra layer to say `OK, you've seen your credit report from the CRAs. You kind of know what the bank is doing from this tool to say what's actually going to be influencing and what's not.' And then you've got clear kind of guidance or text to say that this is actually what's causing the impact on your credit score or the decision making.''\\ ``Yeah, the information you presented it might help in some regards in terms of `oh, actually I’ve misstated this!' ''\\ ``That is a lot more education and a lot more transparent.''\end{tabular}\\
		RQ9  & Explanations must increase the engagement with the product             & ``I think you get a lot more engagement and as a result I think you could put forward a very plausible case.''                                                                                                                                                                                                                                                                                                                                                                                                                                                                                                                                                                                                                                                                                                                                                                                                                  \\
		RQ10 & Explanations must assist in the transparency obligations               & ``That is a lot more education and a lot more transparent.''                                                                                                                                                                                                                                                                                                                                                                                                                                                                                                                                                                                                                                                                                                                                                                                                                                                                    \\
		RQ11 & Explanations must respect intellectual property/trade secrets          & \begin{tabular}[c]{@{}p{.6\textwidth}@{}}``I think in detail, no, I think that you know that that would be bad for the lenders {[}to reveal too much{]}. {[}\ldots{]} you know it is a little bit like asking Coca Cola to print the formula on the outside of every bottle. I mean it just kind of, you know, you kind of break the model.''\\ ``I mean {[}name{]}, that would be fantastic, but I mean, I guess finance houses will be reluctant to share their lending policy rules. Would they?''\end{tabular}                                                                                                                                                                                                                                                                                                                                                                                                                                 \\
		RQ12 & Explanations must be fit for purpose                                   & ``I think that actually even though it’s a lot longer, it gives a lot more information.''                                                                                                                                                                                                                                                                                                                                                                                                                                                                                                                                                                                                                                                                                                                                                                                                                                       \\ \bottomrule
	\end{tabular}
	\caption{Example excerpts from the pilot studies.}\label{tab:examples}
\end{table}

\end{document}